\documentclass[journal]{IEEEtran}

\usepackage{cite}
\usepackage{graphicx} 
\usepackage{import} 
\usepackage{amsmath,amsfonts}
\usepackage{algorithm}
\usepackage{algpseudocode}
\usepackage{units} 
\usepackage{subcaption} 
\usepackage{mathtools} 
\usepackage{hyperref} 
\usepackage{cleveref} 
\usepackage{color} 
\usepackage{balance}          
\usepackage{array} 

\algnewcommand\Not{\textbf{not}}

\renewcommand{\paragraph}[1]{\noindent\textbf{#1}.}

\DeclareMathOperator*{\argmax}{arg\,max}  
\renewcommand{\vec}[1]{\ensuremath{\mathbf{#1}}} %

\title{BayesCPF: Enabling Collective Perception in Robot Swarms with Degrading Sensors}

\author{Khai Yi Chin and Carlo Pinciroli%
\thanks{K.~Y.~Chin and C.~Pinciroli are with the NEST Lab at Worcester Polytechnic Institute, Worcester, MA 01609, USA (email: kchin@wpi.edu, cpinciroli@wpi.edu).}}

\date{} 

\begin{document}

\maketitle

\begin{abstract}
    The collective perception problem---where a group of robots perceives its surroundings and comes to a consensus on an environmental state---is a fundamental problem in swarm robotics. Past works studying collective perception use either an entire robot swarm with perfect sensing or a swarm with only a handful of malfunctioning members. A related study proposed an algorithm that does account for \emph{an entire swarm of unreliable robots} but assumes that the sensor faults are known and remain constant over time. To that end, we build on that study by proposing the \emph{Bayes Collective Perception Filter (BayesCPF)} that enables robots with continuously degrading sensors to accurately estimate the fill ratio---the rate at which an environmental feature occurs. Our main contribution is the Extended Kalman Filter within the BayesCPF, which helps swarm robots calibrate for their time-varying sensor degradation. We validate our method across different degradation models, initial conditions, and environments in simulated and physical experiments. Our findings show that, regardless of degradation model assumptions, fill ratio estimation using the BayesCPF is competitive to the case if the true sensor accuracy is known, especially when assumptions regarding the model and initial sensor accuracy levels are preserved.
\end{abstract}
\section{Introduction}

Perception is a core problem in robot autonomy. The ability of a robot to perceive its surroundings is foundational to its ability to navigate and complete assigned tasks. Our work studies the \emph{collective perception} problem, where we use a swarm of robots to collectively observe and make decisions about the environment in which they operate. Notable examples of such decision-making applications are object classification~\cite{majcherczykDistributedDataStorage2021,stegagnoObjectRecognitionSwarm2014}, infrastructure health assessment~\cite{siemensmaCollectiveBayesianDecisionmaking2024}, and detection of environmental pollutant sources~\cite{tamjidiUnifyingConsensusCovariance2021}. Importantly, environmental conditions are unpredictable in their effect on the sensors used to observe the said environment. This, in addition to the inherent limitations of the sensors due to noise or faults, implies that the measurements collected by the robots can be unreliable. Note that we use the term ``sensor'' as an abstraction of a more general detection mechanism, which can include a variety of binary outcome models, from image classifiers to analog transducers with an ``on/off'' threshold.

\begin{figure}[t]
    \centering
    \def\svgwidth{\columnwidth}
    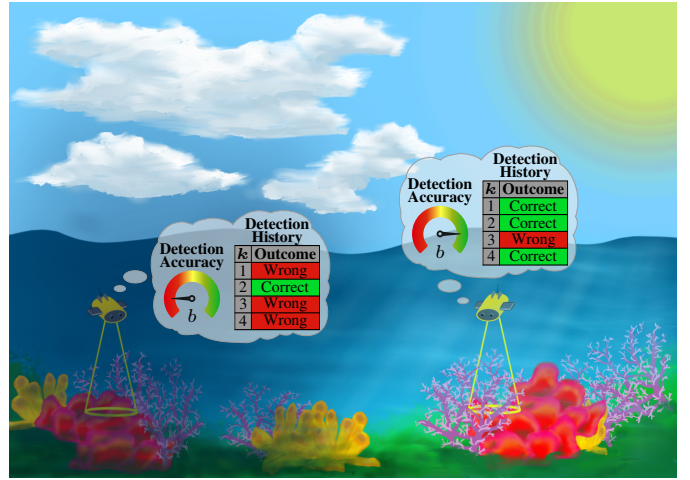
    \caption{An example application where swarm robots are deployed to collectively assess the bleaching coverage of coral reefs. Image classification models (that identify the presence of bleaching) onboard the robots may be less accurate due to suboptimal lighting conditions, as experienced by the robot on the left. Accordingly, it makes more detection mistakes due to a worse detection accuracy $b$.}
    \label{fig:banner_image}
\end{figure}

As a motivating example, we illustrate in \Cref{fig:banner_image} the scenario in which a swarm of underwater robots is deployed to assess the coverage of coral bleaching. Crucial for understanding ecological impacts, the monitoring of coral reefs using image classification methods has been widely studied~\cite{gleasonAutomatedClassificationUnderwater2007,xuDetectionCoralReef2021,gonzalez-riveroMonitoringCoralReefs2020,borbonCoralHealthIdentification2021}. The use of robots can alleviate the labor-intensive process of collecting image data and streamline the assessment of bleaching coverage with onboard sensing. However, depending on geographical features and weather conditions, unpredictable underwater lighting may negatively affect the nominal detection accuracies ascribed to the classification models. Accordingly, we investigate the situation where the detection accuracy of swarm robots degrades over time.

Inspired by Valentini \textit{et al.}~\cite{valentiniCollectivePerceptionEnvironmental2016}, we consider an arena of black and white tiles parametrized by a fill ratio $f$, using black tiles to represent the presence of a bleached coral colony. Each robot is equipped with a ground-facing sensor to observe whether the surface beneath it is black. Over time, the detection accuracy deteriorates, slowly rendering the robots' observations increasingly unreliable. We model the deterioration rate using the Wiener process with drift---a common modeling choice for sensor degradation behavior~\cite{zhangDegradationDataAnalysis2018,hachemDifferentMethodsRUL2024}. For this work, we develop a model robust to continuous sensor degradation with a monotonic expectation. The problem we focus on is akin to a scenario in which underwater light intensity reduction is induced by a moving cloud cover over the region of interest or by a change in the sun's position.

Our research answers the question: ``Can a swarm of robots with imperfect sensors perform collective perception accurately if their sensors worsen over time?'' To be precise, the question contains two interconnected parts: the original problem of \emph{estimating the environment fill ratio} and the ancillary problem of \emph{estimating the sensor accuracy used in solving the original problem}. While past studies on the collective perception problem are extensive, few have studied it using robots with degrading sensors.

We address that gap by introducing an Extended Kalman Filter (EKF) for the ancillary problem, which computes an \emph{estimated sensor accuracy to help the robots self-calibrate} and subsequently solves the original problem. Because the measurement model in our EKF relies on the solution to the first problem, we integrate the EKF with a previously developed algorithm~\cite{chinMinimalisticCollectivePerception2023}---which we refer to as the Minimalistic Collective Perception (MCP) algorithm in this paper. We call this combined algorithm the Bayes Collective Perception Filter (BayesCPF), which---alongside auxiliary functionalities---forms our overarching approach. The parametrization of the two internal components (which we detail in \Cref{sec:methodology} as standalone modules) is performed recursively: the EKF computes sensor accuracy estimates that parametrize the MCP algorithm, which in turn provides fill ratio estimates that parametrize the EKF. The computational requirements of the BayesCPF are also minimal, with constant time and space complexities with respect to the swarm size, supporting its usage on swarm robots---which are usually computationally constrained.

We evaluate the effectiveness of the BayesCPF by running experiments to test various environments and model parameters. In general, the BayesCPF excels in estimating the fill ratio with pre-calibrated sensors and is robust to model inaccuracies. The ambiguity of the environment also plays a significant role in the estimation performance: the BayesCPF works better in an unambiguous environment---where, for example, black tiles appear more frequently than white tiles---as opposed to an ambiguous one---where black and white tiles appear almost as frequently. Our experimentation on physical swarm robots shows that the BayesCPF is effective.

The remainder of the paper is structured as follows: \Cref{sec:related_work} discusses the literature pertaining to collective perception with degrading sensors. \Cref{sec:problem_statement} formulates the problem that we solve in this work, while \Cref{sec:methodology} describes our approach in detail. We review the performance of the BayesCPF in \Cref{sec:simulated_experiments,sec:physical_experiments} before concluding in \Cref{sec:conclusion}.
\section{Related Work} \label{sec:related_work}
Collective perception represents a large class of problems. A commonly studied variant in the swarm robotics literature is the best-of-$n$ problem. The best-of-$n$ problem is one where robots collectively decide the \emph{best} of multiple options; within the collective perception context, the robots' observations of the environment are used to facilitate that decision. An example is the site-selection problem studied by Schmickl \textit{et al.}~\cite{schmicklCollectivePerceptionRobot2007}. They designed a honeybee-inspired approach for a robot swarm to pick the best of two sites. Another honey bee-inspired approach is proposed by Talamali \textit{et al.}~\cite{talamaliImprovingCollectiveDecision2019}, whose heuristic method is sensitive to temporal effects: agents interact with each other increasingly often the longer the elapsed duration. A seminal work that investigates a best-of-$2$ problem is by Valentini \textit{et al.}~\cite{valentiniCollectivePerceptionEnvironmental2016}. They consider a black-and-white tile environment---which we use in this work---where a swarm of robots must collectively decide the color that appears most frequently using a heuristic algorithm called the \emph{voter model}. Later works on the best-of-$n$ problem include data-driven~\cite{almansooriComparativeStudyDecision2022} and probabilistic approaches~\cite{ebertBayesBotsCollective2020,shanDiscreteCollectiveEstimation2021,shanManyoptionCollectiveDecision2024,abdelliMaximumLikelihoodEstimate2023}. Our research is similar to these works in that our robots compute an estimate that can be converted into a best-of-$n$ decision (on whether the environment has more black tiles than white), though we do not show that here. Nonetheless, the quality of our robots' estimates reflects the quality of the decision if converted.

The research mentioned thus far considers an idealistic scenario where robots have little to no flaws in their sensors. Works that \emph{do} take flawed robots into account often include only a subset of malfunctioning robots or involve considerable computational costs. Crosscombe \textit{et al.}~\cite{crosscombeRobustDistributedDecisionmaking2017} introduced a three-state extension to Valentini's voter model to counteract against a subset of unreliable agents while blockchain-based methods proposed by Strobel \textit{et al.}~\cite{strobelManagingByzantineRobots2018,strobelBlockchainTechnologySecures2020,strobelRobotSwarmsNeutralize2023} improve decision-making robustness in the presence of malicious robots, but are computationally demanding. Algorithms by Crosscombe and Lawry~\cite{crosscombeCollectivePreferenceLearning2021} (rank learning) and by Morlino \textit{et al.}~\cite{morlinoCollectivePerceptionSwarm2010} (evolutionary) do account for noisy measurements received by \emph{all the robots}, although communication is global in those works; we consider only robots with local communication. Chin \textit{et al.}~\cite{chinMinimalisticCollectivePerception2023} developed the Minimalistic Collective Perception (MCP) algorithm that tolerates an entire swarm of faulty robots, but they formulated their method with the assumption that the robots' sensor flaws are known and do not change over time. We use this previously developed method as the basis for estimating the fill ratio, but instead, our robots \emph{do not} have access to the true sensor accuracy and degrading sensors.

Our main contribution is a recursive Bayes estimator---an Extended Kalman Filter (EKF)---that determines the aforementioned true sensor accuracy. In particular, our approach provides an estimate, defined as the \emph{assumed sensor accuracy}, that we use in place of the true accuracy to predict the fill ratio, akin to a self-calibration scheme. We integrate both the MCP algorithm (for solving the original estimation problem) and the EKF (for solving the ancillary estimation problem) into a broader algorithm, which we call the Bayes Collective Perception Filter (BayesCPF). While using Bayesian estimation to predict sensor degradation is not new, especially in reliability and industrial engineering, these works often consider estimating only the degradation of a single entity~\cite{elwanyRealtimeEstimationMean2009,liuStochasticFilteringApproach2020,hachemDifferentMethodsRUL2024,yeBayesianApproachCondition2015}. In studies where a network of sensors is concerned \cite{fascistaUnifiedBayesianFramework2023,liuNovelAlgorithmQuantized2023}, the approaches rely on a centralized computation node for estimation. By comparison, the robots in our work are part of a \emph{decentralized swarm}, each responsible for its degradation estimation. The robots also move, scattering across the environment, thus having no guaranteed interactions with other members.

The approach we present here parallels the research of \emph{fault detection, diagnosis, and recovery} methods in swarm robots, an area gaining interest in the swarm intelligence community. For fault detection, Christensen \textit{et al.}~\cite{christensenExogenousFaultDetection2007} proposed a neural-network classifier while Tarapore \textit{et al.}~\cite{taraporeFaultDetectionSwarm2019} took a biology-inspired approach based on the natural immune system. Building on the work of Tarapore \textit{et al.}, O'Keeffe \textit{et al.} implemented heuristic and learning approaches to diagnose detected faults~\cite{okeeffeFaultDiagnosisRobot2017,okeeffeFaultDiagnosisRobot2017a,okeeffeAdaptiveOnlineFault2018}. The BayesCPF instead behaves more closely to a \emph{fault recovery} method---as it helps robots calibrate their assumed sensor accuracy---which is less common in the swarm robotics literature. This is unsurprising as recovery strategies are expected to be associated with errors specific to the application and robots involved. In a swarm collective transport problem, Lee and Hauert~\cite{leeDataDrivenMethodIdentify2024} use a data-driven method that enables a faulty robot to self-detect, which is then isolated from the other robots. Azizi \textit{et al.}~\cite{aziziHierarchicalArchitectureCooperative2012} proposed a hierarchical framework that requires non-isolated members to implement recovery procedures for distributed satellite formation. Taking a more general approach, Parker and Kannan~\cite{parkerAdaptiveCausalModels2006} introduced a method that adaptively updates a robot's recovery action based on a set of pre-defined rules. Our approach differs from the aforementioned works in that we calibrate for the degradation levels of a \emph{decentralized robot swarm}---without isolating faulty members---using a probabilistic approach based on first principles.
\section{Problem Statement} \label{sec:problem_statement}
\begin{figure}[t]
    \centering
    \includegraphics[width=0.9\columnwidth]{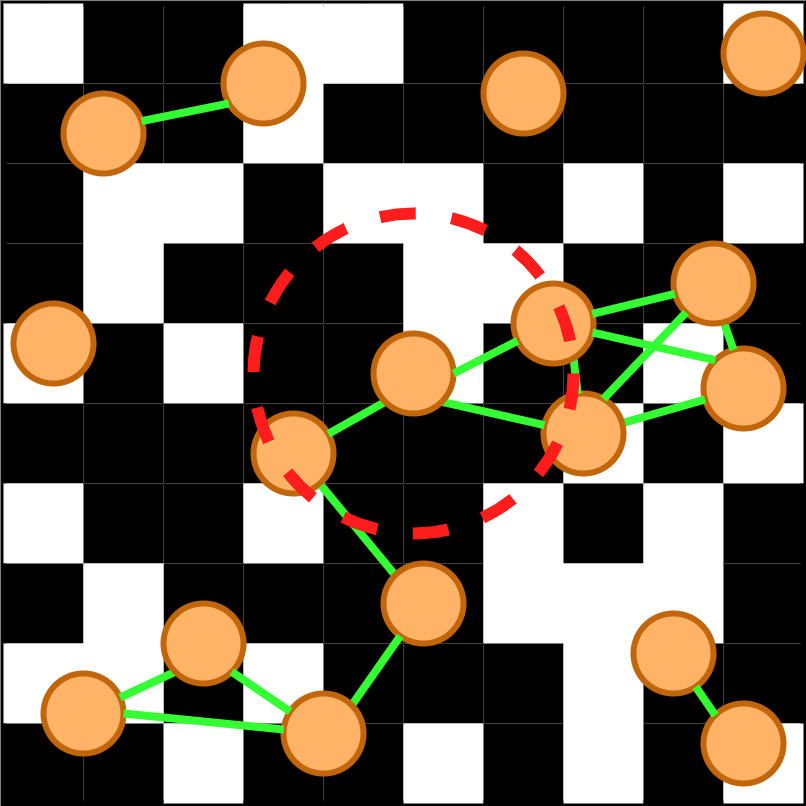}
    \caption{Environment setup for collective perception using a robot swarm. The orange circles indicate moving robots while the green lines indicate communication links. The red dotted line around a robot denotes the communication neighborhood of that robot.}
    \label{fig:mcp_setup}
\end{figure}

We study a problem setup similar to that proposed by Valentini \textit{et al.} \cite{valentiniCollectivePerceptionEnvironmental2016}, visually depicted in \Cref{fig:mcp_setup}. The setup has been studied extensively in past collective perception works~\cite{morlinoCollectivePerceptionSwarm2010,ebertBayesBotsCollective2020,shanDiscreteCollectiveEstimation2021,almansooriComparativeStudyDecision2022,pfisterCollectiveDecisionmakingBayesian2022,pfisterCollectiveDecisionmakingChange2023,abdelliMaximumLikelihoodEstimate2023,siemensmaCollectiveBayesianDecisionmaking2024} and represents an environment where robots observe the occurrence of a feature to make decisions about said environment.

A swarm of $N$ robots is deployed to a square arena of uniformly scattered black or white tiles. The rate at which the black tiles occur in the arena is parametrized by the environment fill ratio, $f \in [0, 1]$. The robots move randomly across this arena---a diffusion process. Each robot collects tile color observations $z_i[k]$ using an onboard ground sensor at time step $k$. The robots' goal is to collectively estimate $f$. In the context of using swarm robots to assess coral bleaching coverage (\Cref{fig:banner_image}), the black tiles represent the \emph{presence} of bleached coral colonies, while the white tiles represent their \emph{absence}. The estimated $f$ thus informs us of the rate of coral bleaching in the environment.

The simplicity of such a setup is key to why it has been widely used in past collective perception studies. In essence, the setup generalizes by abstracting away the occurrence and detection mechanisms specific to any environment or robot, making it applicable to various collective perception tasks. Yet, when studied using swarm robots---which are largely simplistic and homogeneous behaviorally---the setup offers complexity to estimating the environment fill ratio $f$. The challenge is compounded in this work, in that we consider that robots have unreliable sensor measurements. The robots also rely on \textit{decentralized} coordination---a defining trait of swarm systems---and thus, information exchange only happens between a robot and its nearest neighbors without the facilities of centralized computation.

When robot $i$ makes an observation of the tile color $z_i[k]$, it is only sometimes correct; the probability at which this occurs is captured by $b_i[k]$, which yields the following observation probability at any time step $k$:
\begin{equation} \label{eq:observation_prob}
    \begin{aligned}
        p(z_i[k] = 1 \mid \text{black tile encountered}) & = b_i[k],    \\
        p(z_i[k] = 0 \mid \text{black tile encountered}) & = 1 - b_i[k]
    \end{aligned}
\end{equation}
where $z_i[k] = 1$ indicates that robot $i$ observed a black tile and otherwise for $z_i[k] = 0$. The probabilities are flipped for the conditional where a white tile is encountered since the observation has a binary outcome:
\begin{equation*}
    \begin{aligned}
        p(z_i[k] = 1 \mid \text{white tile encountered}) & = 1 - b_i[k], \\
        p(z_i[k] = 0 \mid \text{white tile encountered}) & = b_i[k].
    \end{aligned}
\end{equation*}

Our work combines the \emph{collective perception} problem with the \emph{sensor degradation estimation} problem. The collective perception problem can be formalized as a centralized problem as follows:
\begin{equation}\label{eq:mcp_cenproblem}
    \begin{aligned}
        \argmax_{\vec{\hat{x}}[k]} \quad & \prod_i p(\hat{x}_i[k] \mid \vec{z}_i[k]; b_i[k]) \\
        \text{s.t.}        \quad         & \forall i \; \hat{x}_i[k] = \bar{x}[k]
    \end{aligned}
\end{equation}
where---at time step $k$---$\hat{x}_i[k]$ is each robot's estimate of the state of the world, $\vec{\hat{x}}[k] = [ \hat{x}_1[k], \dots, \hat{x}_N[k] ]^T$, and $\bar{x}[k]$ is the collective estimate. $\vec{z}_i[k]$ is each robot's imperfect sensor observations over a time period $\Delta k$ up until time step $k$.

Instead, our contribution expands this problem to estimate the sensor accuracies as part of a larger estimation problem
\begin{align} \label{eq:sensor_deg_problem}
    \begin{aligned}
        \argmax_{\vec{\hat{x}}[k], \vec{\hat{b}}[k]} \quad & \prod_i p(\hat{x}_i[k], \hat{b}_i[k] \mid \vec{z}_i[k]) \\
        \text{s.t.}        \quad                           & \forall i \; \hat{x}_i[k] = \bar{x}[k]
    \end{aligned}
\end{align}
where $\vec{\hat{b}}[k] = [ \hat{b}_1[k], \dots, \hat{b}_N[k] ]^T$.
\section{Methodology} \label{sec:methodology}

\begin{figure*}[t]
    \centering
    \def\svgwidth{\textwidth}
    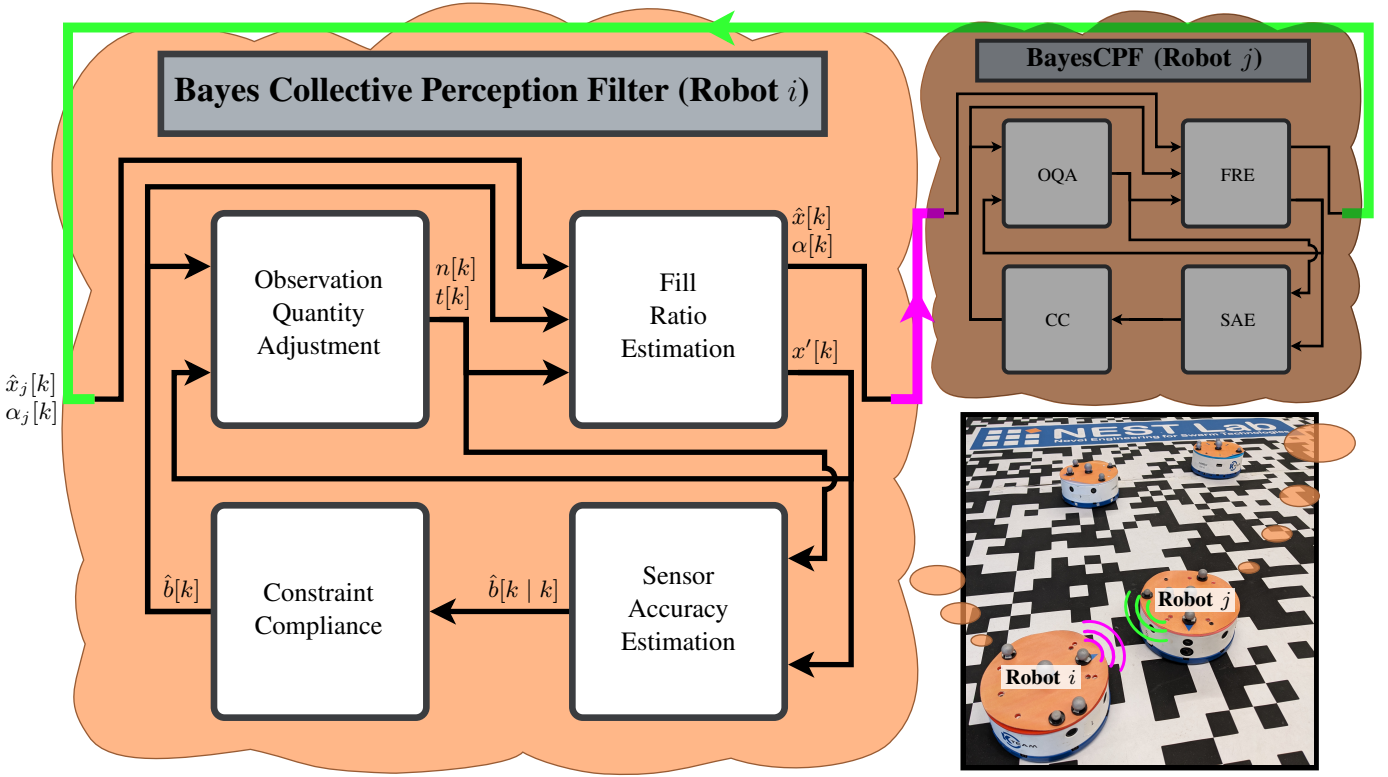
    \caption{The flow of information in the Bayes Collective Perception Filter with the focus on robot $i$. Robot $i$ shares its \emph{local estimate} (of the environment fill ratio $f$) $\hat{x}_i[k]$ with its neighbor, robot $j$, as indicated by pink lines. It also receives robot $j$'s local estimate, $\hat{x}_j[k]$ (green lines). Internally, robot $i$ uses the received $\hat{x}_j[k]$ to compute a \emph{weighted moving average informed estimate} $x_i^\prime[k]$ that it uses to determine the (constrained) \emph{assumed sensor accuracy} $\hat{b}_i[k]$. $\hat{b}_i[k]$ in turn influences the observations that robot $i$ use in estimating the fill ratio and sensor accuracy.}
    \label{fig:bayes_cpf_block_diagram}
\end{figure*}

\begin{algorithm}[t]
    \footnotesize
    \hsize=\columnwidth
    \caption{Bayes Collective Perception Filter executed by an individual robot}\label{alg:bayes_cpf}
    \begin{algorithmic}
        \Require $0 < \hat{b} < 1$ \Comment{sensor accuracies}
        \State $t \gets 0$ \Comment{total tiles observed}
        \State $n \gets 0$ \Comment{total black tiles observed}
        \State $x^\prime \gets 0$ \Comment{weighted moving average informed estimate}
        \While{\Not{} $\Call{Done}{}$}
        \State $\Call{Move}{\phantom{}}$
        \If{$T_{obs}$ steps have passed}
        \State $z \gets \Call{ObserveEnvironment}{\phantom{}}$
        \State Insert $z$ into $\mathbf{\Omega}$ \Comment{add observation to queue}
        \EndIf
        \If{$T_{comms}$ steps have passed}
        \State $\mathbf{H} \gets \Call{DetectNeighbors}{\phantom{}}$ \Comment{generate list of neighbors}
        \State $\Call{Broadcast}{\mathbf{H}, \hat{x}, \alpha}$ \Comment{communicate with neighbors}
        \State $(\mathbf{\hat{x}_H}, \mathbf{\alpha_H}) \gets \Call{Receive}{\mathbf{H}}$
        \EndIf
        \If{new observations made}
        \State $(\hat{x}, \alpha, x^\prime) \gets \Call{FillRatioEstimation}{n, t, \hat{b}, \mathbf{\hat{x}_H}, \mathbf{\alpha_H}}$ \Comment{sec.~\ref{sec:fre_module}}
        \EndIf
        \If{$T_{BCPF}$ steps have passed}
        \State $\hat{b}_{k \mid k} \gets \Call{SensorAccuracyEstimation}{n, t, x^\prime}$ \Comment{sec.~\ref{sec:sae_module}}
        \State $\hat{b} \gets \Call{ComputeConstrainedEstimate}{\hat{b}_{k \mid k}}$ \Comment{eq.~\ref{sec:cc_module}}
        \State $(n, t) \gets \Call{ObservationQuantityAdjustment}{\hat{b}, x^\prime}$ \Comment{sec.~\ref{sec:oqa_module}}
        \EndIf
        \EndWhile
    \end{algorithmic}
\end{algorithm}


The Minimalistic Collective Perception (MCP) algorithm was shown by the original authors to be effective in enabling accurate estimation of $f$ in spite of flawed sensors, though the degree of sensor faults needs to be calibrated beforehand. In the event that the faults worsen over time, estimation performance is likely to suffer. The mitigation strategy---which is the essence of this work---is to provide the robots a means to self-calibrate their sensor accuracy. To preserve the minimalistic requirements of the MCP algorithm, we are interested in using observations collected by the robots (without additional data collection) in this self-calibration process.

Our proposed approach, the Bayes Collective Perception Filter (BayesCPF), tackles the original estimation problem---\emph{fill ratio estimation}---and the ancillary estimation problem---\emph{sensor accuracy estimation}---recursively, as shown in \Cref{fig:bayes_cpf_block_diagram}. At any time $k$, the BayesCPF addresses \Cref{eq:sensor_deg_problem} recursively by solving the following decomposed problems:
\begin{align}
    \hat{x}_i[k] & = \argmax_{{x}_i} p({x}_i \mid \vec{z}_i[k], \hat{b}_i[k - 1]), \label{eq:decomp_fre_problem} \\
    \hat{b}_i[k] & = \argmax_{{b}_i} p({b}_i \mid \vec{z}_i[k], \hat{x}_i[k]). \label{eq:decomp_sae_problem}
\end{align}
Each swarm robot runs its copy of the BayesCPF, which consists of four modules. The overall algorithm is shown in \Cref{alg:bayes_cpf}.

\subsection{Fill Ratio Estimation (FRE) Module} \label{sec:fre_module}
The original estimation problem concerns the ability of robots with imperfect sensors to make decisions about the environment. The Fill Ratio Estimation (FRE) module addresses this by solving \Cref{eq:decomp_fre_problem} using the MCP algorithm, which we describe next.

From \Cref{eq:observation_prob}, we model each tile color observation as a Bernoulli trial. Then, given a robot $i$'s true sensor accuracy $b_i[k]$ at time step $k$, the probability that it sees a black tile is
\begin{align}\label{eq:success_prob}
    \begin{split}
        p(z_i[k] = 1) & = b_i[k] f + (1 - b_i[k])(1 - f) \\
                      & \eqcolon q_i[k]
    \end{split}
\end{align}
where $f$ is the environment's fill ratio. The sum of independent Bernoulli trials is a Binomial random variable, which we use to describe the probability of robot $i$ observing $n_i[k]$ black tiles in $t[k]$ observations (at the $k$\textsuperscript{th} time step):
\begin{multline}\label{eq:binom_likelihood}
    p \bigg( \sum_{l=0}^{t[k]-1} z_i[k - l] = n_i[k] \bigg) \\
    = {t[k] \choose n_i[k]} q_i^{n_i[k]}[k] (1 - q_i[k])^{t[k]-n_i[k]}
\end{multline}
where the success probability $q_i[k]$ is given in \Cref{eq:success_prob}. Note that the sum over $z_i[k-l]$ collects the past $t[k]$\footnote{In this work, we assume the robots are synchronous, hence the absence of the subscript $i$ for the observations made so far.} observations. If we set $t[k]$ to the number of observations made thus far, the summation accumulates the entire observation history. Using the entire observation history can be counterproductive, however, as older observations may be generated from a distribution different from the recent ones due to sensor degradation. We designed the Observation Quantity Adjustment (OQA) module (\Cref{sec:oqa_module}) specifically to address this issue.

As detailed in the original MCP work, by using the black-tile-observation probability as our likelihood function, we find robot $i$'s \textit{local estimate} $\hat{x}_i[k]$ as the maximum likelihood estimator of $f$:
\begin{equation}\label{eq:local_estimate}
    \hat{x}_i[k] =
    \begin{cases}
        0                                                                                 & \text{if } \frac{\nicefrac{n_i[k]}{t[k]} + \hat{b}_{i}[k - 1] - 1}{2 \hat{b}_{i}[k - 1] - 1} \leq 0, \\[10pt]
        1                                                                                 & \text{if } \frac{\nicefrac{n_i[k]}{t[k]} + \hat{b}_{i}[k - 1] - 1}{2 \hat{b}_{i}[k - 1] - 1} \geq 1, \\[10pt]
        \frac{\nicefrac{n_i[k]}{t[k]} + \hat{b}_{i}[k - 1] - 1}{2 \hat{b}_{i}[k - 1] - 1} & \text{otherwise}.
    \end{cases}
\end{equation}
We associate a \textit{local confidence} $\alpha_i[k]$ to this estimate, found using the estimator's Fisher information:
\begin{equation}\label{eq:local_confidence}
    \alpha_i[k] =
    \begin{cases}
        \frac{\nu^2 \big( t[k]\hat{b}_{i}^2[k] - \nu (t[k] - n_i[k]) \big)}{\hat{b}_{i}^2[k](\hat{b}_{i}[k - 1]-1)^2} & \text{if } \hat{x}_i[k] = 0, \\[10pt]
        \frac{\nu^2 \big( t[k]\hat{b}_{i}^2[k] - \nu n_i[k] \big)}{\hat{b}_{i}^2[k] (\hat{b}_{i}[k - 1] - 1)^2}       & \text{if } \hat{x}_i[k] = 1, \\[10pt]
        \frac{\nu^2  t^3[k]}{n_i[k](t[k] - n_i[k])}                                                                   & \text{otherwise}
    \end{cases}
\end{equation}
where $\nu \coloneq 2\hat{b}_{i}[k - 1] - 1$. The use of $\hat{b}_i[k - 1]$ here indicates that the \emph{assumed} sensor accuracy is used in estimating the fill ratio, as robot $i$ does not have access to $b_i[k]$.

Then, the robots $j$ communicate their local values $\hat{x}_j[k]$ and $\alpha_j[k]$ with neighboring robots to exchange information. For each recipient robot $i$, these values are combined into a \textit{social estimate} $\bar{x}_i[k]$ and \textit{social confidence} $\beta_i[k]$:
\begin{align}
    \bar{x}_i[k] & = \frac{1}{\sum_{j \in \mathcal{J}_i[k]} \alpha_j[k]} \sum_{j \in \mathcal{J}_i[k]} \alpha_j[k] \hat{x}_j[k], \label{eq:social_estimate} \\
    \beta_i[k]   & = \sum_{j \in \mathcal{J}_i[k]} \alpha_j[k]. \label{eq:social_confidence}
\end{align}
where $\mathcal{J}_i[k]$ is the set of robot $i$'s neighbors at time step $k$.

Finally, robot $i$ consolidates both its and its neighbors' information with the \textit{informed estimate} $x_i[k]$:
\begin{align}\label{eq:informed_estimate}
    \begin{split}
        x_i[k] & = \frac{\alpha_i[k] \hat{x}_i[k] + \beta_i[k] \bar{x}_i[k]}{\alpha_i[k] + \beta_i[k]}                                                                \\[4pt]
               & = \frac{\alpha_i[k] \hat{x}_i[k] + \sum_{j \in \mathcal{J}_i[k]} \alpha_j[k] \hat{x}_j[k]}{\alpha_i[k] + \sum_{j \in \mathcal{J}_i[k]} \alpha_j[k]}.
    \end{split}
\end{align}
Notably, this weighted average of the local and social estimates---a result inspired by decentralized Kalman filtering---is robot $i$'s best guess of the environment fill ratio $f$.

Additionally, for uses in \Cref{sec:sae_module,sec:oqa_module}, we propose that robot $i$ tracks a weighted moving average of past $Q \in \mathbb{Z^+}$ informed estimates
\begin{align}\label{eq:weighted_informed_estimate}
    x_i^\prime[k] & = \frac{x_i[k] + 2 x_i[k-1] + \ldots + Q x_i[k-(Q-1)]}{1 + 2 + \ldots + Q}.
\end{align}
Note that \Cref{eq:weighted_informed_estimate} is part of our contribution for this work and is separate from the original MCP algorithm. Our emphasis on \textit{older} informed estimates (larger weights on older informed estimates) reflects an assumption we make about robotic field applications. We expect the robots to be well calibrated before being deployed; thus, earlier informed estimates (made while knowledge of the sensor accuracy level is quite precise) should be more reliable than later ones. $x_i^\prime[k]$ is an internal quantity used by the BayesCPF and is not used to report fill ratio estimation performance.

\begin{algorithm}[t]
    \footnotesize
    \hsize=\columnwidth
    \caption{Fill Ratio Estimation Module}\label{alg:fre_module}
    \begin{algorithmic}
        \Function{FillRatioEstimation}{$n ,t, \hat{b}, \mathbf{\hat{x}_H}, \mathbf{\alpha_H}$}
        \State $(\hat{x}, \alpha) \gets \Call{LocalEstimation}{n, t, \hat{b}}$ \Comment{eqs.~\ref{eq:local_estimate}, \ref{eq:local_confidence}}
        \If{$\mathbf{\hat{x}_H} \neq$ empty $\And \mathbf{\alpha_H} \neq$ empty } \Comment{check list of neighbors}
        \State $(\bar{x}, \beta) \gets \Call{SocialEstimation}{\mathbf{\hat{x}_H}, \mathbf{\alpha_H}}$ \Comment{eqs.~\ref{eq:social_estimate}, \ref{eq:social_confidence}}
        \Else
        \State $(\bar{x}, \beta) \gets (0, 0)$
        \EndIf
        \State $x \gets \Call{InformedEstimation}{\hat{x}, \alpha, \bar{x}, \beta}$ \Comment{eq.~\ref{eq:informed_estimate}}
        \State $x^\prime \gets \Call{WeightedMovingAverage}{x}$ \Comment{eq.~\ref{eq:weighted_informed_estimate}}
        \State \Return $x^\prime$
        \EndFunction
    \end{algorithmic}
\end{algorithm}

\subsection{Sensor Accuracy Estimation (SAE) Module}\label{sec:sae_module}
The FRE module relies on the assumed sensor accuracy $\hat{b}_i[k]$ to estimate the environment fill ratio $f$. Because the robots' sensor worsens over time, we track the time-varying sensor accuracy (\Cref{eq:decomp_sae_problem}) using recursive Bayes estimation, which is central to our Sensor Accuracy Estimation (SAE) module. We stress here that the modules in the current and subsequent subsections are original work and constitute our main contribution for this paper.

We implement a popular form of the recursive Bayes estimator: the Extended Kalman Filter (EKF), which assumes the underlying prior and likelihood models to be approximately Gaussian \cite{thrunProbabilisticRobotics2005}. More precisely, the prior is the integral of the product between the \textit{state transition model} and the \textit{posterior of the previous time step}, both of which are assumed to be approximately Gaussian. These models are used in the EKF in two sequential steps, \textit{predict} and \textit{update}.

However, due to the Gaussian assumption, the EKF is designed for state variables with a $(-\infty, \infty)$ support. Practical sensor accuracy ranges, on the other hand, are finite, complicating the use of the EKF to track sensor degradation. To that end, we extended the SAE module with a PDF truncation method \cite{simonConstrainedKalmanFiltering2010} that satisfies the linear inequality constraints imposed on state variables, which we describe in \Cref{sec:cc_module}. We omit the subscript $i$ in the equations hereafter for brevity as they apply to robots individually. (They remain in \Cref{tab:symbol_definitions} to ensure the symbol definitions are explicit.) Note that the prior and posterior assumed sensor accuracy estimates computed by the EKF below are denoted with a conditional (\textit{e.g.,} $\hat{b}[k \mid k - 1]$ and $\hat{b}[k \mid k]$). These are \emph{not} the assumed sensor accuracy $\hat{b}[k]$.

\subsubsection{Predict step}
This step leverages domain expertise (manifested in the state transition model) to predict the state variables' evolution. For the state transition model, we consider the degradation of the sensor accuracy to be a Wiener process. The true sensor accuracy for a robot evolves across time as follows:
\begin{align} \label{eq:wiener_process}
    b[k] = b[k-1] + \eta + \gamma G
\end{align}
where $\eta$ is the drift coefficient, $\gamma$ is the diffusion coefficient, and $G$ is the standard Brownian motion (which follows a standard Gaussian distribution):
\begin{align*}
    G \sim \mathcal{N}(0, 1).
\end{align*}
In this work, we assume that the true values of $\eta$ and $\gamma$ may not always be available during initialization; thus, we parametrize our algorithm with $\tilde{\eta}$ and $\tilde{\gamma}$ as our best knowledge of said parameters.

As such, each increment $\Delta b[k] = b[k] - b[k-1]$ is an i.i.d.\ Gaussian random variable with a mean of $\tilde{\eta}$ and a variance of $\tilde{\gamma}^2$. Then, we obtain the predicted assumed sensor accuracy $\hat{b}[k \mid k-1]$ and the predicted variance $\rho^2[k \mid k-1]$ as
\begin{align} \label{eq:ekf_predict}
    \hat{b}[k \mid k-1] & = \hat{b}[k-1 \mid k-1] + \tilde{\eta}    \\
    \rho^2[k \mid k-1]  & = \rho^2[k-1 \mid k-1] + \tilde{\gamma}^2
\end{align}
where $\hat{b}[k-1 \mid k-1]$ and $\rho^2[k-1 \mid k-1]$ are the previous assumed sensor accuracy estimate and estimate variance, respectively. This work considers non-positive values for $\tilde{\eta}$ to mirror real-world sensors.

As for the initial posterior, we assume that the true sensor accuracy $b[k]$ is Gaussian between each time step (the linear inequality constraints will be satisfied after the \textit{update} step).

\subsubsection{Update step}
This step leverages observations of the state variables to correct the model-based predictions, given a measurement likelihood. Although our likelihood provided in \Cref{eq:binom_likelihood} is Binomial, its limiting distribution is Gaussian:
\begin{align}
    \lim_{t[k]\rightarrow\infty}{n[k]} \sim \mathcal{N} \big(\mu, \sigma^2 \big)
\end{align}
where
\begin{align*}
    \mu      & = t[k] q[k],           \\
    \sigma^2 & = t[k] q[k](1 - q[k]).
\end{align*}
That is, the observation that a robot makes has an expected value of $t[k] q[k]$, which we define as the measurement model
\begin{align}
    \begin{split}
        h(b[k]) & \coloneq t[k] q[k]              \\
                & = t[k] (b[k]f + (1-b[k])(1-f))  \\
                & = t[k] (b[k] (2f - 1) - f + 1).
    \end{split}
\end{align}

However, the measurement model depends on an unknown quantity: the environment fill ratio $f$---our main estimation target. We circumvent this issue using the robot's informed estimate as a \textit{fill ratio reference}. To prevent instability issues driven by non-stationary estimates, we use the weighted moving average informed estimate $x^\prime[k]$ from \Cref{eq:weighted_informed_estimate} as a quasi-static ``snapshot'' of the fill ratio reference, akin to the \textit{fixed Q-Target} technique in Deep Q-Learning \cite{mnihHumanlevelControlDeep2015}. The updated measurement model is thus
\begin{align}
    h(b[k]) & = t[k] (b[k] (2x^\prime[k] - 1) - x^\prime[k] + 1)
\end{align}
and its Jacobian is
\begin{align}
    \begin{split}
        H[k] & = \frac{\partial}{\partial b[k]} h(b[k])   \\
             & = t[k] \frac{\partial}{\partial b[k]} q[k] \\
             & = t[k] (2x^\prime[k] - 1).
    \end{split}
\end{align}
With this, we find the posterior estimate and variance using the EKF \textit{update} equations:
\begin{align} \label{eq:ekf_update}
    \hat{b}[k \mid k] & = \hat{b}[k \mid k-1] + K[k] (n[k] - h(\hat{b}[k \mid k-1])), \\
    \rho^2[k \mid k]  & = (1 - K[k] H[k]) \rho^2[k \mid k-1]
\end{align}
where the Kalman gain is
\begin{align}
    K[k] & = \frac{\rho^2[k \mid k-1] H[k]}{H^2[k] \rho^2[k \mid k-1] + t[k] q[k](1 - q[k])}.
\end{align}

Due to the asymptotic dependency of the Gaussian approximation, we foresee poor measurement update performance when the number of observations $t[k]$ is small. Accordingly, we impose a rule of thumb condition where the \textit{update} step will be executed only if $t[k]\hat{b}[k \mid k-1] \geq 5$ and $t[k](1-\hat{b}[k \mid k-1]) \geq 5$. This rule is inspired by a widely used rule of thumb when approximating Binomial distributions~\cite{walpoleProbabilityStatisticsEngineers1998}. Otherwise, the predicted values will be assigned as the posterior values instead, \textit{i.e.,} $\hat{b}[k \mid k] \coloneq \hat{b}[k \mid k-1]$ and $\rho^2[k \mid k] \coloneq \rho^2[k \mid k-1]$.


\begin{algorithm}[t]
    \footnotesize
    \hsize=\columnwidth
    \caption{Sensor Accuracy Estimation Module}\label{alg:sae_module}
    \begin{algorithmic}
        \Require $\hat{b}_{0 \mid 0}, \rho^2_{0 \mid 0}$ \Comment{initial guesses for estimate and variance}
        \Require $\tilde{\eta}, \tilde{\gamma}$ \Comment{state transition model parameters}
        \Function{SensorAccuracyEstimation}{$n, t, x^\prime$}
        \State $(\hat{b}_{k \mid k-1}, \rho^2_{k \mid k-1}) \gets \Call{Predict}{\hat{b}_{k-1 \mid k-1}, \rho^2_{k-1 \mid k-1}}$ \Comment{eq.~\ref{eq:ekf_predict}}
        \If{$t\hat{b}_{k \mid k-1} \geq 5 \And t(1-\hat{b}_{k \mid k-1}) \geq 5$}
        \State $(\hat{b}_{k \mid k}, \rho^2_{k \mid k}) \gets \Call{Update}{\hat{b}_{k \mid k-1}, \rho^2_{k \mid k-1}, n, t, x^\prime}$ \Comment{eq.~\ref{eq:ekf_update}}
        \Else
        \State $(\hat{b}_{k \mid k}, \rho^2_{k \mid k}) \gets (\hat{b}_{k \mid k-1}, \rho^2_{k \mid k-1})$
        \EndIf
        \State \Return $\hat{b}_{k \mid k}$
        \EndFunction
    \end{algorithmic}
\end{algorithm}

\subsection{Constraint Compliance (CC) Module} \label{sec:cc_module}
Before using the posterior estimate $\hat{b}[k \mid k]$ as a robot's assumed sensor accuracy, we check for constraint violations using the Constraint Compliance (CC) module. If out-of-bounds, the module alters a copy of the posterior estimate via a truncation method proposed by Simon and Simon~\cite{simonConstrainedKalmanFiltering2010} before assigning it as the assumed sensor accuracy $\hat{b}[k]$. Due to the non-recursive nature of this method, the \textit{original posterior estimate} is fed back as input to the SAE module rather than the constrained assumed sensor accuracy. Here, we outline the equations used only to constrain the $1$-dimensional posterior estimate to ensure that this manuscript is self-contained; the general form of the truncation method can be found in the original work.

Given a constant, linear inequality constraint on a robot's sensor accuracy $c_{l} \leq b[k] \leq c_{u}$, the assumed sensor accuracy output by this module is
\begin{equation} \label{eq:constrained_state_estimate}
    \hat{b}[k] =
    \begin{cases}
        \hat{b}[k \mid k]                       & \text{if } c_{l} \leq b[k] \leq c_{u}, \\
        \rho[k \mid k] y[k] + \hat{b}[k \mid k] & \text{otherwise}.
    \end{cases}
\end{equation}
For clarification, we emphasize the distinction between the $\hat{b}[k \mid k]$ and $\hat{b}[k]$ variables: the former is the posterior value obtained from and used in the SAE module, while the latter is the assumed sensor accuracy that the robot uses in the FRE and OQA modules. The term $y[k]$ is the expected value of a truncated standard Gaussian random variable
\begin{align}
    \begin{split}
        y[k] & = \int^{d_{u}[k]}_{d_{l}[k]} \xi \exp \bigg\{ \frac{-\xi^2}{2} \bigg\} d\xi                                                                                                                                                  \\[4pt]
             & = \frac{\sqrt{\nicefrac{2}{\pi}} \big( \exp \{ \nicefrac{-d_{l}[k]^2}{2} \} - \exp \{ \nicefrac{-d_{u}[k]^2}{2} \} \big)}{\mathrm{erf}\{ \nicefrac{d_{u}[k]}{\sqrt{2}} \} - \mathrm{erf}\{ \nicefrac{d_{l}[k]}{\sqrt{2}} \}}
    \end{split}
\end{align}
where the truncation limits are the original bounds, standardized:
\begin{align}
    d_{l}[k] & = \frac{c_{l} - \hat{b}[k \mid k]}{\rho[k \mid k]}, \\
    d_{u}[k] & = \frac{c_{u} - \hat{b}[k \mid k]}{\rho[k \mid k]}.
\end{align}



\begin{algorithm}[t]
    \footnotesize
    \hsize=\columnwidth
    \caption{Observation Quantity Adjustment Module}\label{alg:oqa_module}
    \begin{algorithmic}
        \Require $|\Omega| \leq Q_{max}$ \Comment{observation queue}
        \Function{ObservationQuantityAdjustment}{$\hat{b}, x^\prime$}
        \State $\dot{\hat{b}} \gets \Call{ComputeTimeDerivative}{\hat{b}}$ \Comment{eq.~\ref{eq:assumed_accuracy_derivative}}
        \State $t \gets \Call{ComputeObservationQuantity}{x^\prime, \dot{\hat{b}}}$ \Comment{eq.~\ref{eq:observation_quantity_final}}
        \State \Return $(n, t)$
        \EndFunction
    \end{algorithmic}
\end{algorithm}

\subsection{Observation Quantity Adjustment (OQA) Module} \label{sec:oqa_module}
In the SAE module, we noted the incompatibility of using the most recent informed estimate as the fill ratio reference due to its non-stationary nature. Likewise, because a robot's observations of tile colors depend on its true sensor accuracy, the distribution of $n[k]$ varies as $b[k]$ changes. This suggests that only recent observations should be considered in calculating $n[k]$---as opposed to accumulating it from the start of a robot's operation---which we achieve with the Observation Quantity Adjustment (OQA) module.

Given an observation queue $\Omega$ with a capacity of $Q_{max}$, our goal is to use a subset of observations from the \emph{same distribution}. We achieve that by examining how quickly the black tile observation rate changes over time, which we find using the time derivative of the success probability $q[k] \coloneq p(z[k] = 1)$ (from \Cref{eq:success_prob}):
\begin{align} \label{eq:time_derivative_success_prob}
    \begin{split}
        \frac{dq[k]}{dk} & = \frac{d}{dk} \big \{ b[k]f + (1-b[k])(1-f) \big \} \\
                         & = (2f - 1) \frac{db[k]}{dk}.
    \end{split}
\end{align}
Naturally, the number of observations $t[k]$ should grow inversely to the rate of change of the distribution that generates the observations, which we enforce with the following rule:
\begin{align} \label{eq:observation_quantity}
    t[k] & =  \left \lceil \frac{\phi}{| T_{obs} (2x^\prime[k] - 1)\dot{\hat{b}}[k] |} \right \rceil
\end{align}
where we incorporate \Cref{eq:time_derivative_success_prob} in the denominator. $\phi$ is the distribution sensitivity level, a tunable parameter that reflects the degree of distribution changes tolerable by the robot operator, and $T_{obs}$ is the observation period. $\phi$ effectively lets the operator adjust the history of observations that should be accounted for in the other modules. It is non-negative and should be tuned in proportion to the rate of sensor accuracy degradation $\eta$ (manifested in $\nicefrac{db[k]}{dk}$). For example, suppose \Cref{eq:time_derivative_success_prob} yields $-1 \times 10^{-4}$. If we define $Z_{100}$ to be the number of observations made in $100$ time steps, this means that the probability $q[k]$ is expected to decrease by $0.01$ every $Z_{100} T_{obs} = 100$ time steps. In this case, the operator would set $\phi = 1$ if they find the $0.01$ decline to be acceptable, $\phi < 1$ if it is too much, and $\phi > 1$ if it is too little.

For stability reasons, $x^\prime[k]$ is used in \Cref{eq:observation_quantity} as the fill ratio reference, as is done in the SAE module. This reasoning is extended to our use of $\dot{\hat{b}}[k]$ in place of $\nicefrac{db[k]}{dk}$, where---aside from using the \emph{assumed sensor accuracy} as a proxy for the true sensor accuracy---we compute a moving average of a robot's past \emph{assumed accuracy degradation rate} with a fixed window size $W$:
\begin{align} \label{eq:assumed_accuracy_derivative}
    \dot{\hat{b}}[k] & \coloneq \frac{\frac{d\hat{b}[k]}{dk} + \frac{d\hat{b}[k-1]}{dk} + \ldots + \frac{d\hat{b}[k-(W-1)]}{dk}}{1 + 2 + \ldots + W}
\end{align}
and we calculate the time difference of $\hat{b}[k]$ for the derivative: $\nicefrac{d\hat{b}[k]}{dk} \coloneq \hat{b}[k] - \hat{b}[k-1]$.

We also use $t[k]$ in deciding how many past informed estimates to use in the computation of $x^\prime[k]$ from \Cref{eq:weighted_informed_estimate} (in place of $Q$). One may notice a circular dependence between \Cref{eq:weighted_informed_estimate,eq:observation_quantity}: calculating $x^\prime[k]$ requires $t[k]$, which in turn needs $x^\prime[k]$ for its calculation. We avoid this by only executing \Cref{eq:observation_quantity} when enough iterations of the algorithm have run to compute $\dot{\hat{b}}[k]$. Therefore, the number of observations $t[k]$ used in \Cref{eq:local_estimate,eq:local_confidence,eq:local_confidence,eq:weighted_informed_estimate} is
\begin{align} \label{eq:observation_quantity_final}
    t[k] & =
    \begin{cases}
        \min \bigg \{ Q_{max}, \left \lceil \frac{\phi}{| T_{obs} (2x^\prime[k] - 1)\dot{\hat{b}}[k] |} \right \rceil \bigg \} & \text{if } k \geq W T_{obs}, \\[10pt]
        \min \bigg \{ Q_{max}, \left \lfloor \frac{k}{T_{obs}} \right \rfloor \bigg \}                                         & \text{otherwise}
    \end{cases}
\end{align}
where we cap the quantity to a maximum threshold $Q_{max}$ to accommodate the typically limited resources on swarm robots. We reiterate that the queue capacity remains at $Q_{max}$ at all times; thus, $t[k]$ corresponds to the \emph{size of a subset} of $\Omega$ that contains the most recent in-distribution observations.

\begin{table}[t]
    \centering
    \caption{Definitions of recurring symbols. Symbols without the subscript $i$ imply that they apply identically to all robots in the swarm. Symbols without the time step index $[k]$ imply that they remain constant throughout the robot swarm's operation.}
    \renewcommand{\arraystretch}{1.5} 
    \begin{tabular}{>{\centering}m{1.75cm} m{6cm}}
        \hline
        \textbf{Symbol}                            & \textbf{Definition}                                                                \\
        \hline
        $f$                                        & Environment fill ratio                                                             \\
        $b_i[k]$                                   & True sensor accuracy of robot $i$ at time step $k$                                 \\
        $\hat{b}_i[k]$                             & Assumed sensor accuracy of robot $i$ at time step $k$                              \\
        $z_i[k]$                                   & Observation of robot $i$ at time step $k$                                          \\
        $p(z_i[k] = 1)$ \newline $\coloneq q_i[k]$ & Black tile observation probability of robot $i$ at time step $k$                   \\
        $n_i[k]$                                   & Number of black tiles seen by robot $i$ at time step $k$                           \\
        $t[k]$                                     & Number of tiles seen by each robot at time step $k$                                \\
        $\hat{x}_i[k]$                             & Local estimate of robot $i$ at time step $k$                                       \\
        $x_i[k]$                                   & Informed estimate of robot $i$ at time step $k$                                    \\
        $x^\prime_i[k]$                            & Weighted moving average of $x_i[k]$                                                \\
        $\eta$                                     & True drift coefficient of the Wiener model                                         \\
        $\tilde{\eta}$                             & Assumed drift coefficient of the Wiener model                                      \\
        $\gamma$                                   & True diffusion coefficient of the Wiener model                                     \\
        $\tilde{\gamma}$                           & Assumed diffusion coefficient of the Wiener model                                  \\
        $\phi$                                     & Distribution sensitivity level used to determine the number of observations to use \\
        $W$                                        & Window size used to compute $\dot{\hat{b}}_i[k]$                                   \\
        $Q_{max}$                                  & Capacity of the observation queue                                                  \\
        $T_{obs}$                                  & Observation period (in time steps)                                                 \\
        $T_{comms}$                                & Communication period (in time steps)                                               \\
        $T_{BCPF}$                                 & Sensor accuracy update period (in time steps)                                      \\
        \hline
    \end{tabular}
    \label{tab:symbol_definitions}
\end{table}

\subsection{Remarks}

\begin{figure}[t]
    \centering
    \def\svgwidth{\columnwidth}
    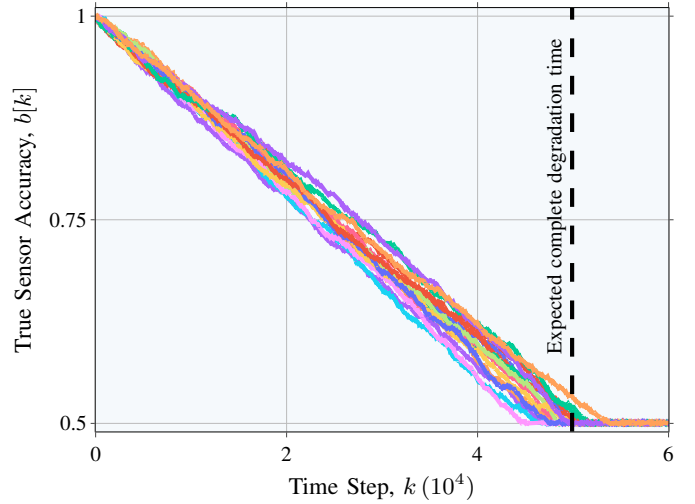
    \caption{An example of true sensor accuracy degradation behavior as governed by the Wiener process (\Cref{eq:wiener_process}) with $\eta = -1 \times 10^{-5}$ and $\gamma = 1 \times 10^{-4}$, each curve representing one of 15 robots. The lowest possible accuracy level is fixed at 0.5 as this is the point where the sensor returns a completely random observation.}
    \label{fig:true_sensor_acc_example}
\end{figure}

\paragraph{Sensor Accuracy Range}
Because the sensor accuracy $b[k]$ is an abstraction of the accuracy of a binary classifier, we set the range of its possible values as $0.5 \leq b[k] \leq 1.0$ in this work. The use of $0.5$ as our lower bound is due to the fact that a $50\%$ accurate binary classifier has truly random outcomes and is \emph{least} useful than those at any other accuracy level. We refer to an instance where the sensor accuracy level falls to $0.5$ as \emph{catastrophic sensor degradation}. In the experimental evaluation sections, we simulated this degradation behavior by saturating the true and assumed sensor accuracy---$b[k]$ and $\hat{b}[k]$ respectively---at $0.5$ or $1.0$, as shown in \Cref{fig:true_sensor_acc_example} (though we do study cases with non-catastrophic sensor degradation as well).

\paragraph{Computational Complexity}
When scaling with respect to swarm size, time, and space complexity is $O(1)$ for the BayesCPF. Aside from the three queues used to store $z_i[k]$, $x_i[k]$, and $\hat{b}_i[k]$ values (whose size can be adjusted based on practical constraints), robot $i$ only stores a single copy of the other variables. The communication bandwidth requirement follows the MCP algorithm, which is $O(1)$. The minimalistic nature of the BayesCPF makes it a viable algorithm for swarm robots, which are often assumed to be limited in onboard computational resources.

\section{Simulated Experiments} \label{sec:simulated_experiments}

\subsection{General Setup and Metrics}

\paragraph{Setup}
Our simulated experiments are conducted using ARGoS, a physics-based, multi-robot simulator \cite{pinciroliARGoSModularParallel2012}. In each of the experiments, we run $M = 30$ trials using $N=15$ robots modeled after the Khepera IV \cite{soaresKheperaIVMobile2016}, as can be seen in the bottom right of \Cref{fig:bayes_cpf_block_diagram}. The Khepera IV robot has a differential drive with four ground-facing sensors that detect whether the surface underneath is black; we only use the front right ground sensor in our experiments. The robot can also detect obstacles and communicate with neighboring robots within a radius $R$; we set $R = \unit[0.7]{m}$---five times the diameter of the robot. For $K_{max} = 60{,}000$ steps (each time step equivalent to $\unit[0.1]{s}$), the robots diffuse across a simulated environment of black-and-white tiles uniformly scattered according to a target fill ratio $f$, collecting observations while avoiding other robots. We consider experimental environments from a spectrum between \textit{ambiguous} to \textit{unambiguous}: $f = \{ 0.55, 0.65, 0.75, 0.85, 0.95 \}$. The swarm density we tested is $D = \nicefrac{N \pi R ^2}{L^2} = 1$, where $L$ is the simulated arena's edge length that we control for a specific $D$. Note that---in this and the subsequent experimental sections---we remove the subscript $i$ from the individual robot variables ($b_i[k], \hat{b}_i[k], n_i[k], \hat{x}_i[k], x_i[k]$)  to declutter our discussion of the BayesCPF performance.

Unless otherwise specified, the true drift coefficient of the robots sensors is $\eta = -1.0 \times 10^{-5}$ (\Cref{fig:true_sensor_acc_example}), while we vary the assumed drift coefficients in the experiments: $\tilde{\eta} = \{ -0.5, -1.0, -1.5 \} \times 10^{-5}$. (The drift coefficient is effectively the expected degradation rate of the sensor accuracy level.) We set the assumed diffusion coefficient to be the same as the true value, $\tilde{\gamma} = \gamma = 1.0 \times 10^{-4}$, to prioritize testing cases where sensors degrade; the \emph{diffusion} coefficient does not affect sensor degradation in its expectation. For the initial sensor accuracies, we test the following values:
\begin{itemize}
    \item Initial true sensor accuracy $b[0] = \{ 1.0, 0.8 \}$
    \item Initial assumed sensor accuracy $\hat{b}[0] = \{ 1.0, 0.8 \}$
\end{itemize}
While the entire swarm has the same degradation parameters ($\eta, \tilde{\eta}, \gamma, \tilde{\gamma}, b[0], \hat{b}[0]$) within a trial, individual robots of said swarm are not likely to have identical levels of sensor accuracy ($b[k], \hat{b}[k]$) at any point in time; the random number generators are not seeded the same across the swarm. We set the distribution sensitivity factor $\phi = 0.02$, the fixed window size $W = 500$, and the observation queue capacity $Q_{max} = 1{,}000$ used in the OQA module (\Cref{eq:observation_quantity,eq:assumed_accuracy_derivative,eq:observation_quantity_final}). The robots observe the environment and communicate at the same rate of $\nicefrac{1}{T_{obs}} = \nicefrac{1}{T_{comms}} = \unit[2]{Hz}$. The FRE module computes the informed estimates at the rate the control loop runs ($\unit[10]{Hz}$) while the sensor accuracy update modules (SAE, CC and OQA) update at $\nicefrac{1}{T_{BCPF}} = \unit[2]{Hz}$.


\begin{figure*}[t]
    \centering
    \def\svgwidth{\textwidth}
    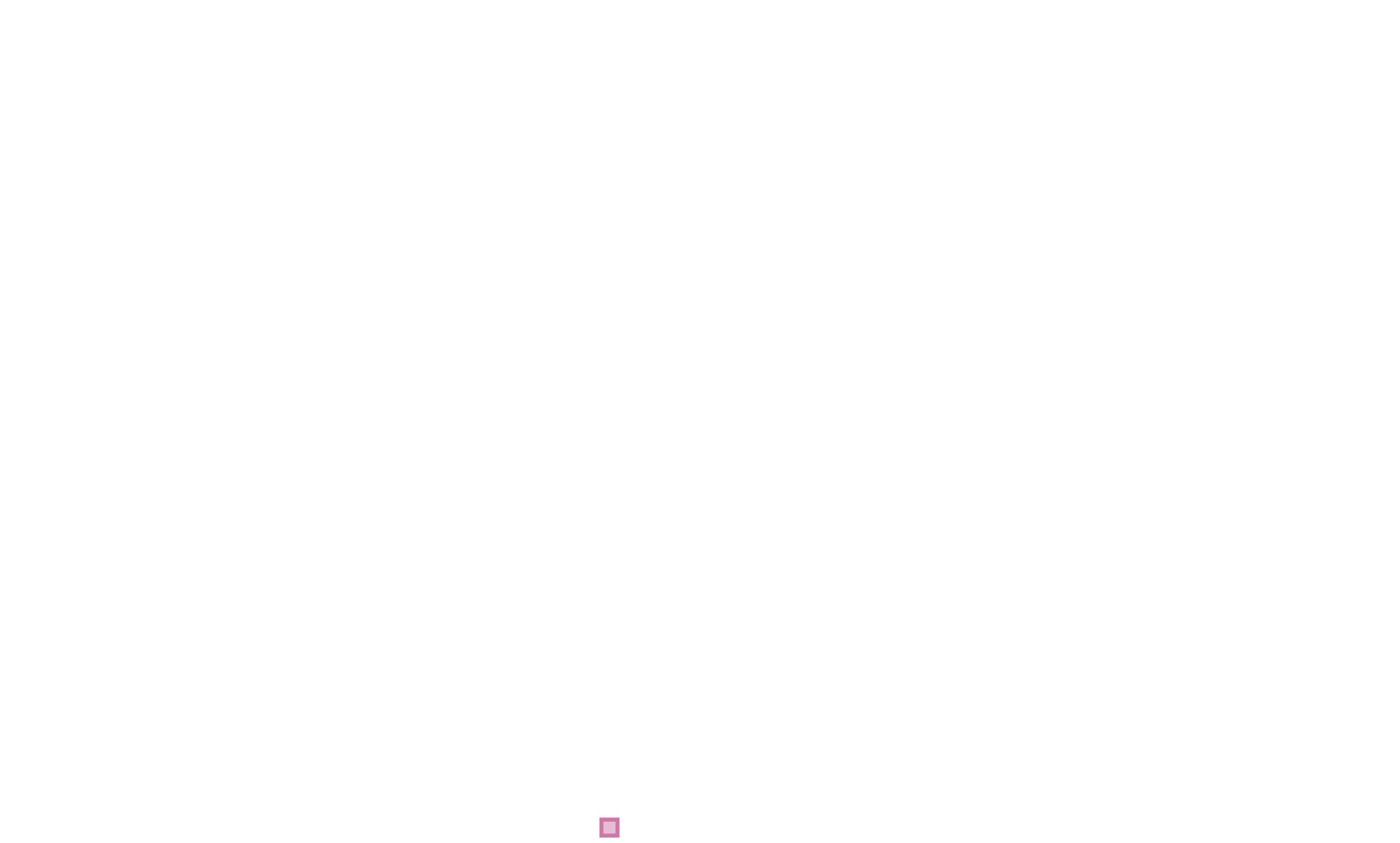
    \caption{\emph{Informed estimation errors} when using the BayesCPF on robots with \emph{catastrophic sensor degradation ($b[k] \rightarrow 0.5$)}; lower values indicate better performance. Each box (and its outliers) contains data points from $M = 30$ trials. The initial true accuracy $b[0]$ is either $1.0$ (left column, secondary $x$-axis) or $0.8$ (right column, secondary $x$-axis). $x[k]$ estimation is better in situations where the modeling assumptions are preserved, \textit{i.e.,} in the transient phase where the state transition model follows true sensor degradation behavior.}
    \label{fig:delta_cumulative_rmsd_regular_inf_est_ldal500}
\end{figure*}

\begin{figure*}[t]
    \centering
    \def\svgwidth{0.975\textwidth}
    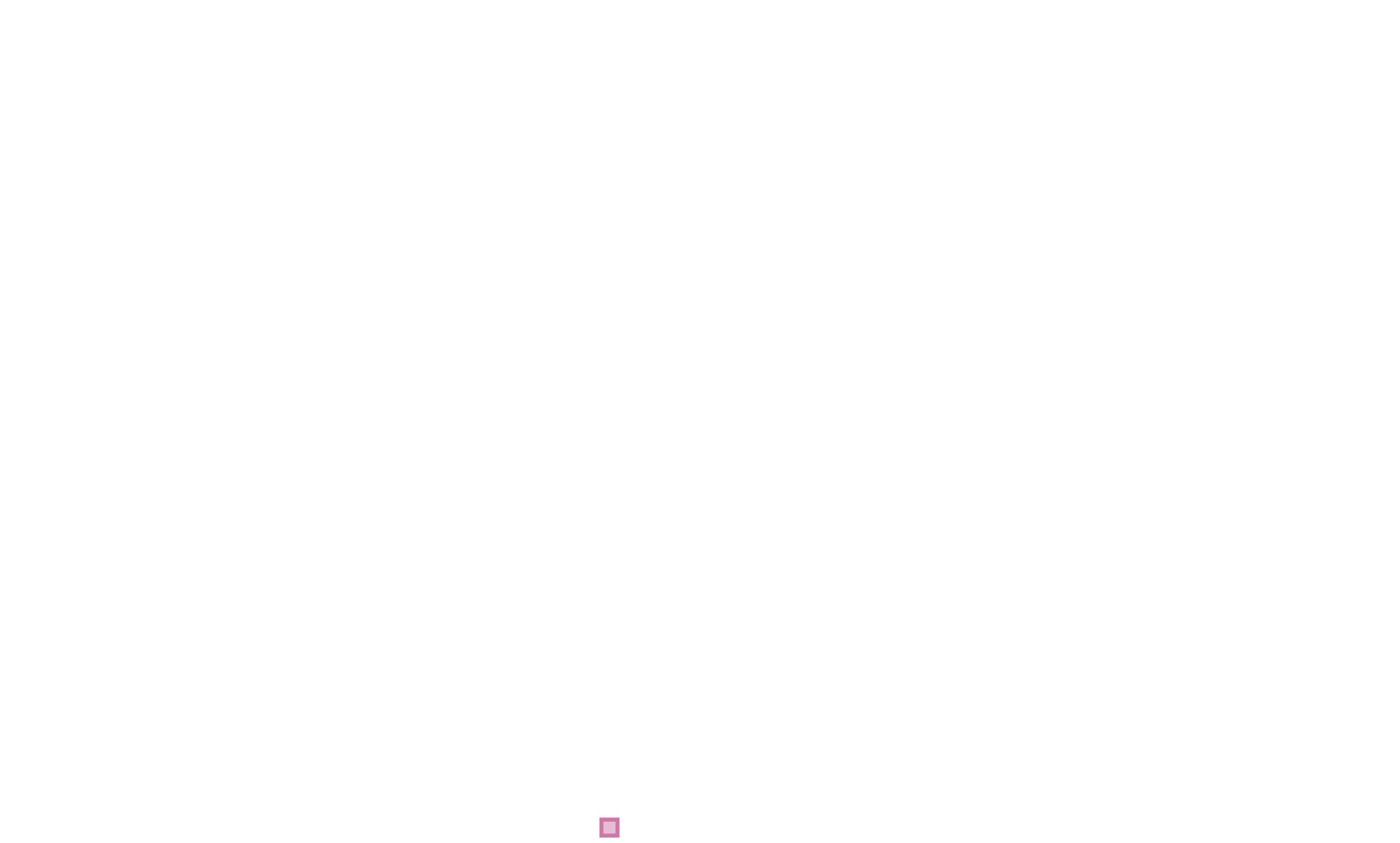
    \caption{\emph{Assumed sensor accuracy estimation errors} when using the BayesCPF on robots with \emph{catastrophic sensor degradation ($b[k] \rightarrow 0.5$)}; lower values indicate better performance. Each box (and its outliers) contains data points from $M = 30$ trials. The initial true accuracy $b[0]$ is either $1.0$ (left column, secondary $x$-axis) or $0.8$ (right column, secondary $x$-axis). $\hat{b}[k]$ estimation is excellent in the equilibrium phase because the BayesCPF has a lower bound of $0.5$.}
    \label{fig:delta_cumulative_rmsd_sensor_acc_ldal500}
\end{figure*}

\begin{figure*}[!t]
    \centering
    \begin{subfigure}{0.32\textwidth}
        \centering
        \def\svgwidth{\textwidth}
        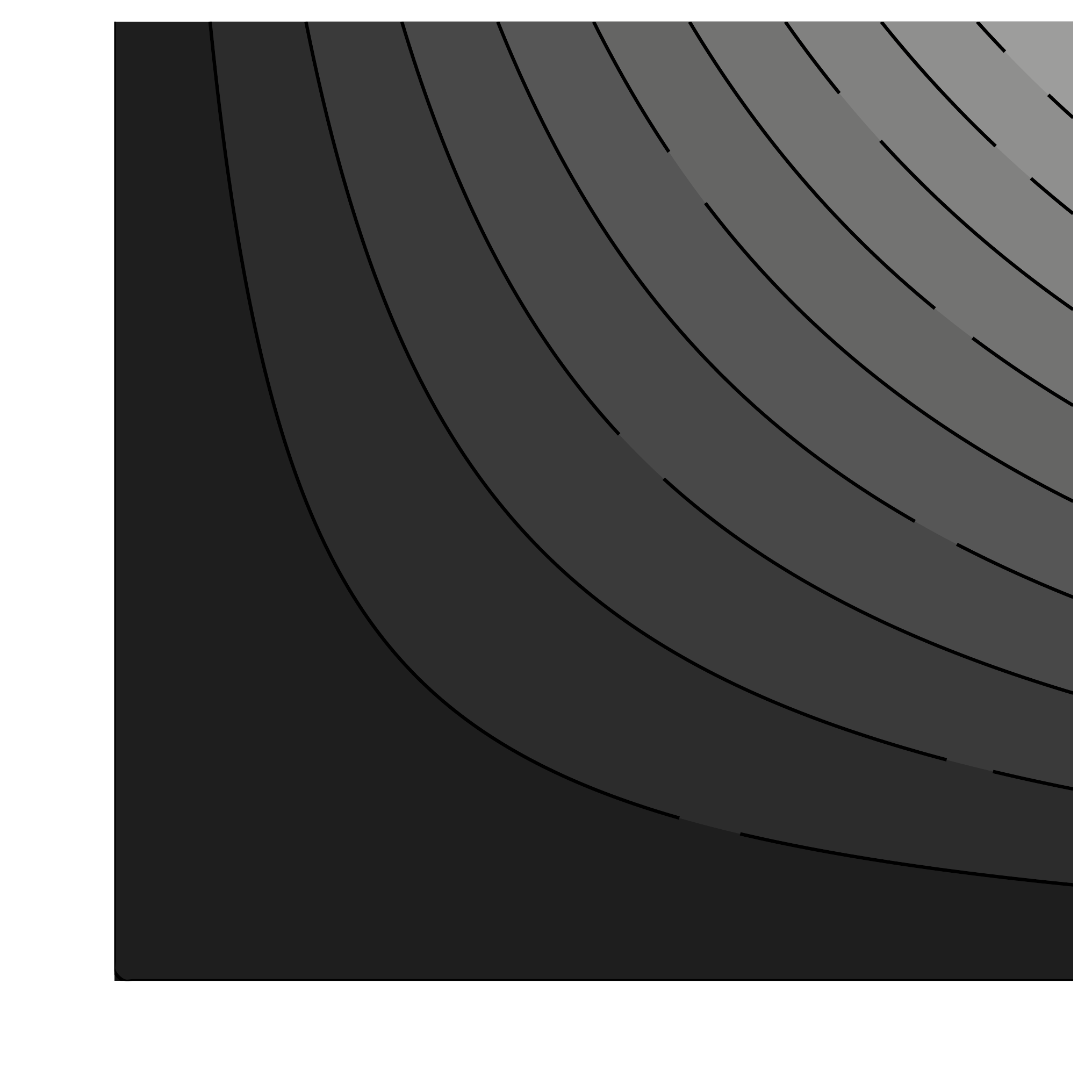
    \end{subfigure}
    \begin{subfigure}{0.32\textwidth}
        \centering
        \def\svgwidth{\textwidth}
        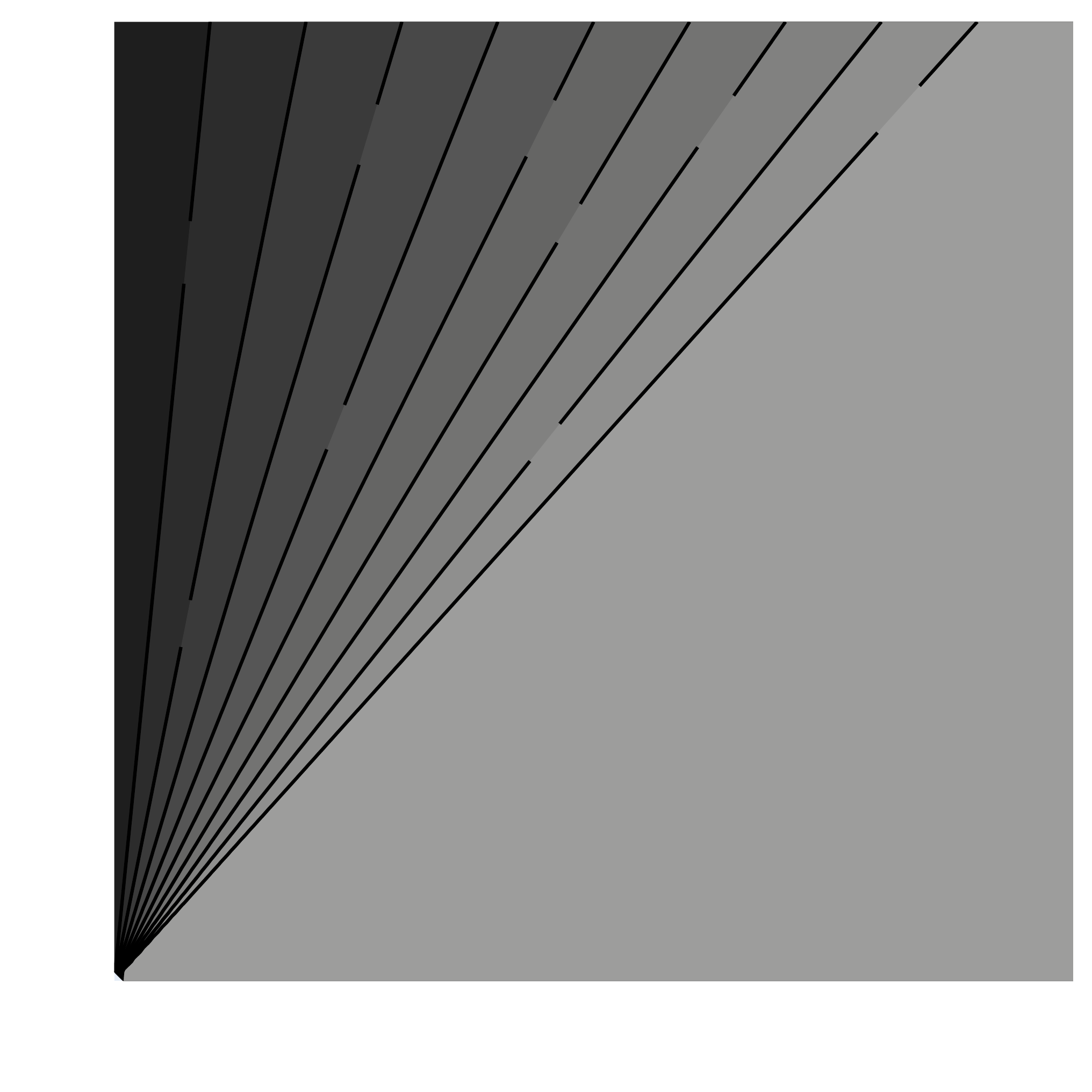
    \end{subfigure}
    \begin{subfigure}{0.32\textwidth}
        \centering
        \def\svgwidth{\textwidth}
        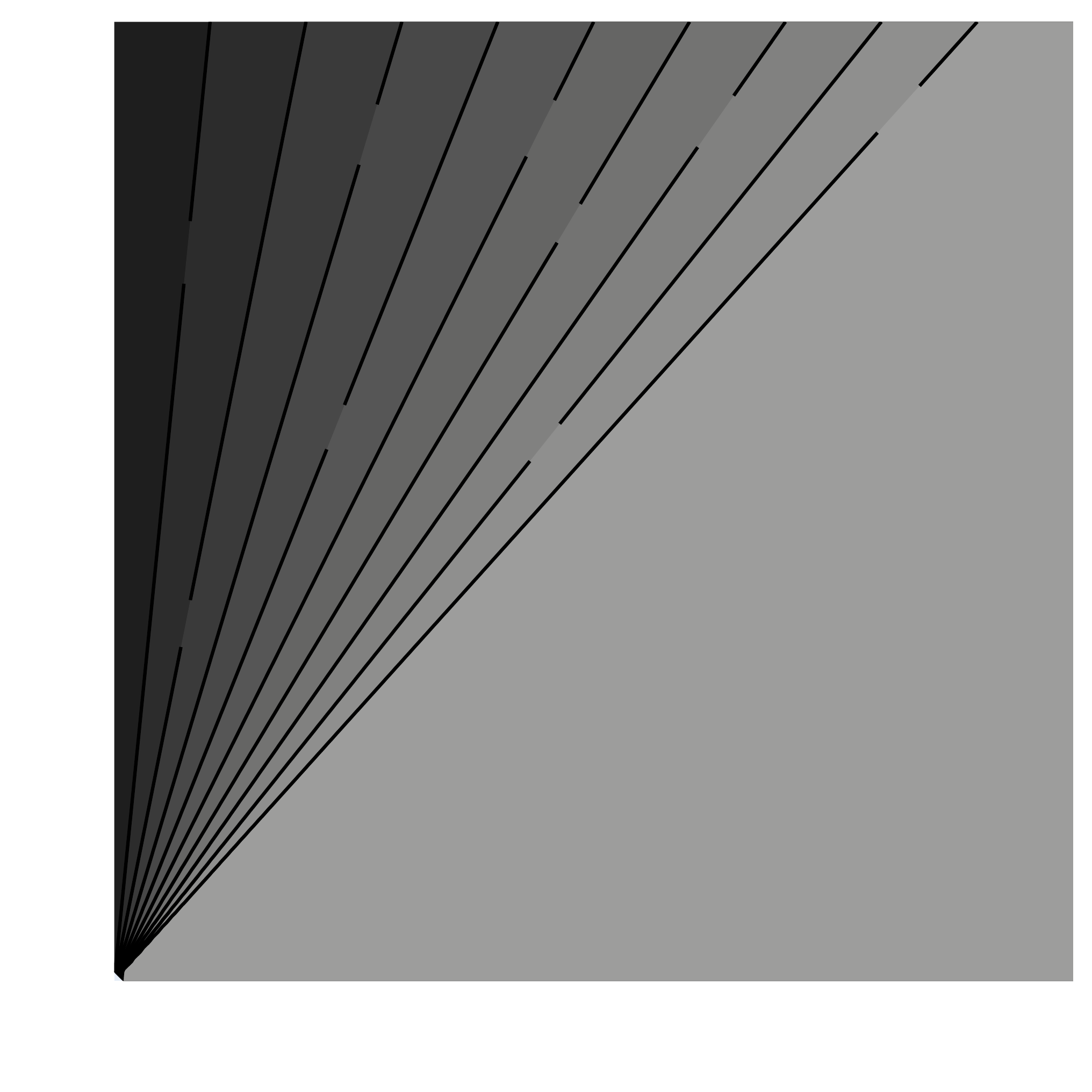
    \end{subfigure}
    \caption{Heatmaps of the \textit{expected black tile observation rate} $\mathbb{E}[\nicefrac{n[k]}{t[k]}]$ (left) and \textit{local estimate} $\hat{x}[k]$ (center) from \Cref{eq:success_prob,eq:local_estimate} respectively. The right heatmap is the center heatmap with the phase portrait of $\hat{x}[k]$ overlaid, assuming that the assumed sensor accuracy $\hat{b}[k]$ evolves linearly to $\nicefrac{n[k]}{t[k]}$. We also assume that $\nicefrac{n[k]}{t[k]}$---acting as a proxy to the true sensor accuracy $b[k]$---decreases deterministically in said heatmap. In reality, due to non-deterministic observations (noisy $\nicefrac{n[k]}{t[k]}$), the trajectory $\hat{x}[k]$ takes is more prone to breaching the contour lines when $\hat{b}[k]$ approaches $0.5$.}
    \label{fig:heatmap}
\end{figure*}

\paragraph{Metrics}
Our metric to evaluate the estimation quality of the BayesCPF is the \emph{root-mean-square deviation} (RMSD). The RMSD enables the aggregation of a swarm's residuals and captures the spread of said residuals---useful for characterizing consensus behavior.

For each trial $m$, we find the informed estimate RMSD at each time step $k$ as
\begin{align*}
    \delta^{[m]}_x[k] = \sqrt{\frac{1}{N}\sum_{i=1}^N (x_i^{[m]}[k] - f)^2}
\end{align*}
where $x_i^{[m]}[k]$ is robot $i$'s informed estimate and $f$ is the environment fill ratio. We compute an analogous metric for the assumed sensor accuracy:
\begin{align*}
    \delta^{[m]}_b[k] = \sqrt{\frac{1}{N}\sum_{i=1}^N (\hat{b}_i^{[m]}[k] - b_i^{[m]}[k])^2}
\end{align*}
where $\hat{b}_i^{[m]}[k]$ is robot $i$'s assumed sensor accuracy and $b_i^{[m]}[k]$ is its true sensor accuracy.

To assess our method's effectiveness under various operating conditions, we further consolidate the metric into a normalized RMSD (N-RMSD) of a single trial $m$ as
\begin{align*}
    \zeta_u^{[m]} \coloneq \frac{1}{\Delta K_\text{phase}}\sum_{k = K_\text{phase\,start}}^{K_\text{phase\,end}} \delta_u^{[m]}[k]
\end{align*}
where $u \in \{b, x\}$ and $\Delta K_\text{phase} = K_\text{phase\,end} - K_\text{phase\,start}$ is the duration of the \textit{sensor degradation phase} of interest.

The different sensor degradation phases refer to the \textit{transient} and the \textit{equilibrium} phases. As the labels suggest, the \textit{transient} phase is the period where a robot's sensor deteriorates according to \Cref{eq:wiener_process}, while the \textit{equilibrium} phase is the stretch that follows where the degradation plateaus. We decide where the phase changes from one to another by calculating the expected degradation, \textit{i.e.,} \Cref{eq:wiener_process} without the diffusion term. An example is shown in \Cref{fig:true_sensor_acc_example}, where the dashed line marks the point of phase change.

%

\subsection{Catastrophic Sensor Degradation ($b[k] \rightarrow 0.5$)} \label{sec:results_ldal500}
First, we analyze the effectiveness of the BayesCPF in situations where the robots' sensor can go from decent to completely unusable, \textit{i.e.,} $b[k] \rightarrow 0.5$. As shown in \Cref{fig:delta_cumulative_rmsd_regular_inf_est_ldal500}, the BayesCPF excels in estimating the environment fill ratio during the \textit{transient} phase but not in the \textit{equilibrium} phase. This is expected as the sensor behavior is consistent with the assumptions of the state transition model in the earlier phase (sensor accuracy is declining) but not in the later phase.

In general, environmental ambiguity plays a substantial role in the effectiveness of the BayesCPF. As we see in \Cref{fig:delta_cumulative_rmsd_sensor_acc_ldal500}, assumed sensor accuracy ($\hat{b}[k]$) estimation performance is worse for ambiguous environments than unambiguous ones, which is as expected. This is corroborated by \Cref{fig:heatmap} (left): at $f=0.55$ (ambiguous), the expected black tile observation rate $\mathbb{E}[\nicefrac{n[k]}{t[k]}]$ is hardly distinguishable for the entire true sensor accuracy range ($0.5 < b[k] < 1.0$).\footnote{In discussing our results, we use the \emph{black tile observation rate} $\nicefrac{n[k]}{t[k]}$ instead of the \emph{number of black tiles seen} $n[k]$ due to the former's finite and constant range: $\nicefrac{n[k]}{t[k]} \in [0, 1]$. The expectation of the rate $\mathbb{E}[\nicefrac{n[k]}{t[k]}]$ is the same as the \emph{black tile observation probability} $p(z[k] = 1)$ used in \Cref{eq:success_prob}. Note that the subscript $i$ is omitted for brevity.} Only with the increase of $f$ do we see the formation of $\nicefrac{n[k]}{t[k]}$ segments favorable for estimation purposes.

\paragraph{Transient phase}
From \Cref{fig:delta_cumulative_rmsd_regular_inf_est_ldal500}, we observe that the informed estimation performance is more sensitive to the \textit{initial assumed sensor accuracy} than it is to the \textit{state transition model}. The N-RMSD remains similar regardless of whether the assumed degradation rate $\tilde{\eta}$ is under-, over-, or accurately estimated, but increases when initial conditions are wrong. This is an encouraging outcome for the BayesCPF as we anticipate the calibration of a robot's initial sensor accuracy to be easier than characterizing its sensor degradation behavior, particularly when the degradation concerns real-world conditions that may be hard to model.

When the initial conditions are correct, the effectiveness of the BayesCPF is largely uniform across different environment fill ratios. Due to a lower resolution of the black tile observation rate $\nicefrac{n[k]}{t[k]}$ in the beginning, a more accurate sensor ($b[0] = 1.0$) reports observations less erroneously, giving us slightly better performance than if sensors are less accurate initially ($b[0] = 0.8$). When the initial conditions are incorrect, estimation performance drops noticeably, though in certain environments, the error remains comparable to those in the experiments with correct initial conditions. Interestingly, we see a discrepancy in the error trends---relative to the fill ratio---between under- and overestimating the assumed sensor accuracy $\hat{b}[0]$. In the former case, estimation performance in unambiguous environments ($f$ closer to $1.0$) exceeds that of ambiguous environments ($f$ closer to $0.5$), but the opposite is true of the latter. We attribute the contrasting behaviors to the combination of environmental and initial conditions, which can be understood through the phase portrait (unrelated to the two sensor degradation phases) of the local estimate $\hat{x}[k]$ in \Cref{fig:heatmap}.

With an \emph{overestimated} $\hat{b}[0]$, the local estimate $\hat{x}[k]$ follows the dotted trajectories in \Cref{fig:heatmap} (right), while the dashed trajectories represent $\hat{x}[k]$ with an \emph{underestimated} $\hat{b}[0]$. We see that the local estimate is likelier to start---and remain---accurate for the \emph{underestimated case when $f = 0.95$} and for the \emph{overestimated case when $f = 0.55$}. In fact, for $f = 0.55$, having a correct starting condition does not contribute to the effectiveness of the BayesCPF as much as it does for $f = 0.95$; all the $\hat{x}[k]$ trajectories at $f = 0.55$ cluster relatively close to its ground truth already. Additionally, because $\nicefrac{n[k]}{t[k]}$ is stochastic in practice, the actual trajectories would include horizontal perturbations. This results in volatile estimates, the degree to which we judge using the distance of a trajectory to its adjacent contour lines; hence, the significantly lower N-RMSD values for $f = 0.95$ when $\hat{b}[0] < b[0]$.

\paragraph{Equilibrium phase}
In this phase, the sensor accuracy for each robot is expected to remain at $b[k] = 0.5$, apart from minor perturbations caused by a zero-mean white noise with variance $\gamma^2$. We see no discernible differences in trend between the various assumed degradation models ($\tilde{\eta}$ \textit{v.s.} $\eta$) and initial starting conditions ($\hat{b}[0]$ \textit{v.s.} $b[0]$). Contrary to the performance in the transient case, however, the BayesCPF unilaterally performs worse in \emph{unambiguous} environments than in \textit{ambiguous} ones.

\begin{figure}[t]
    \centering
    \def\svgwidth{\columnwidth}
    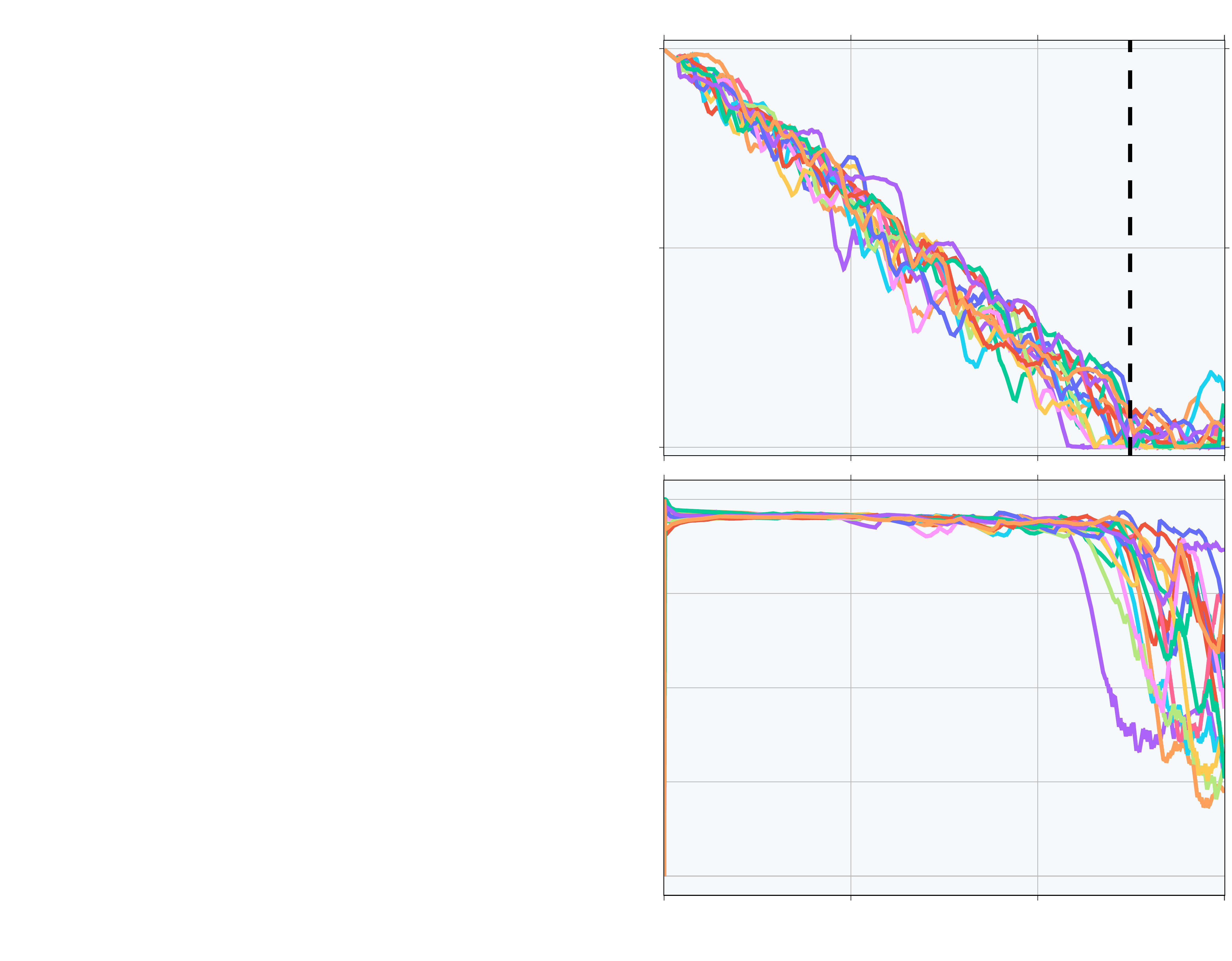
    \caption{Assumed accuracy, $\hat{b}[k]$ and weighted moving average (WMA) informed estimate, $x^\prime[k]$ behaviors for a swarm of 15 robots (15 lines) from a single trial. The robots' assumed degradation rate is $\tilde{\eta} = -1.0 \times 10^{-5}$. The turbulent parts of $x^\prime[k]$---leading up and into the equilibrium phase (past the dashed vertical line)--- highlight the wildly unpredictable informed estimates $x[k]$ that oscillate between $0.0$ and $1.0$.}
    \label{fig:time_series_ldal500}
\end{figure}

The widespread rise in N-RMSD levels is as predicted because the robots' sensors have degraded to a point that information contained in their observations is completely random. This is confirmed using \Cref{eq:success_prob}: the fill ratio $f$ has no bearing on the black tile observation probability $p(z[k] = 1)$ at catastrophic degradation levels ($b[k] = 0.5$)---on average the robot sees $50\%$ of all tiles as black, \textit{i.e.,} $\mathbb{E}[\nicefrac{n[k]}{t[k]}] = 0.5$. Yet, this is inadequate in explaining the \emph{worse estimation performance in unambiguous environments than in ambiguous ones}. The cause can be found with the aid of \Cref{fig:time_series_ldal500}: the weighted moving average (WMA) informed estimate ($x^\prime[k]$) plots show that estimation errors oscillate about 0.5, which is closer to the ambiguous environment ground truth ($f = 0.55$) than it is for the unambiguous one ($f = 0.95$). In essence, the observed trend arises from our evaluation methods rather than any explicit involvement of the BayesCPF.

\begin{figure}
    \centering
    \def\svgwidth{\columnwidth}
    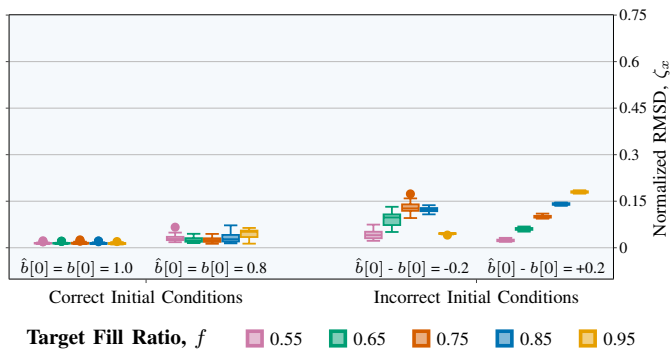
    \caption{\emph{Informed estimation errors} when using the BayesCPF on robots with \emph{no expected sensor degradation} ($\eta = 0$); lower values indicate better performance. Each box (and its outliers) contains data points from $M = 30$ trials. The initial true accuracy $b[0]$ is either $1.0$ (left column, secondary $x$-axis) or $0.8$ (right column, secondary $x$-axis). The results are similar to those in both catastrophic and non-catastrophic sensor degradation (\Cref{fig:delta_cumulative_rmsd_regular_inf_est_ldal500,fig:delta_cumulative_rmsd_regular_inf_est_ldal700}, respectively), underscoring the flexibility of the BayesCPF.}
    \label{fig:cumulative_rmsd_reg_inf_est_zero_drift}
\end{figure}

We also studied, as a baseline, the situation where the robots' sensor accuracy does not degrade (or does so in negligible amounts during their operation). Our results---shown in \Cref{fig:cumulative_rmsd_reg_inf_est_zero_drift}---only include the experiments where our assumed drift coefficient matches the true degradation rate, \textit{i.e.,} $\tilde{\eta} = \eta = 0$. (A zero mean white noise with variance $\gamma^2$ still applies.) An alternative scenario where our assumption of the sensor degradation does not match $\eta = 0$ is just another instance of experiments we discussed earlier and is thus untested.

We see that the performance of the BayesCPF with a non-degrading sensor mirrors that with a degrading sensor. Even the trends are identical: the filter excels when it is initialized correctly and worse when not. This implies the lack of drawbacks to implementing the BayesCPF onboard swarm robots, with the potential upside of the filter providing accurate informed estimates should degradation occur.

Overall, the BayesCPF proves helpful in situations where calibration of the initial assumed sensor accuracy can be achieved. That said, the estimation fares well only during the transient phase but not during the equilibrium phase. This is a reasonable limitation considering that the BayesCPF has relatively low computational demands, which permits supplementary algorithms to improve its estimation deficiencies. In many situations, complete sensor degradation may also take a long time to arrive at to be a practical concern.

\begin{figure*}[t]
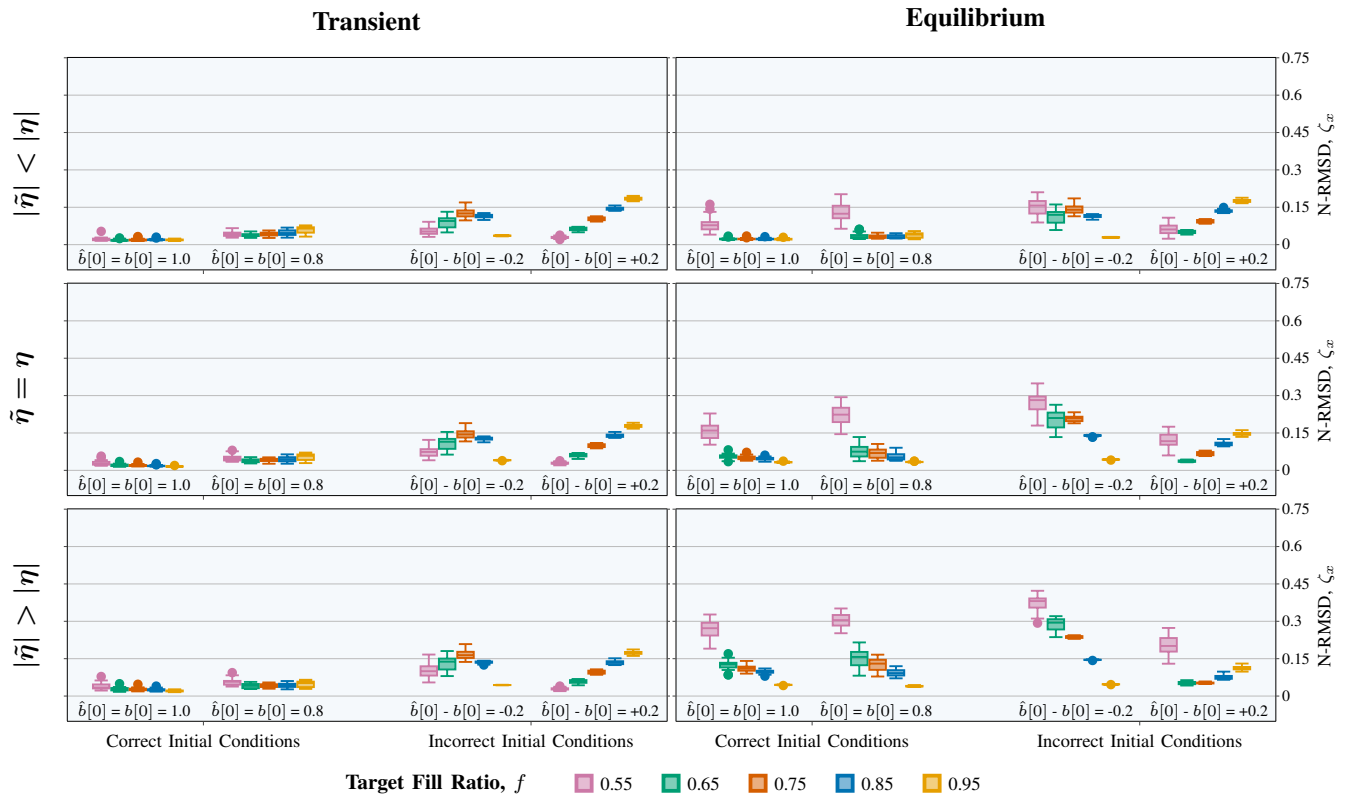

    \centering
    \def\svgwidth{0.975\textwidth}
    
    \caption{\emph{Informed estimation errors} when using the BayesCPF on robots with \emph{non-catastrophic sensor degradation ($b[k] \rightarrow 0.7$)}; lower values indicate better performance. Each box (and its outliers) contains data points from $M = 30$ trials. The initial true accuracy $b[0]$ is either $1.0$ (left column, secondary $x$-axis) or $0.8$ (right column, secondary $x$-axis). The equilibrium phase performance is notably improved over that of the catastrophic sensor degradation case in \Cref{fig:delta_cumulative_rmsd_regular_inf_est_ldal500}.}
    \label{fig:delta_cumulative_rmsd_regular_inf_est_ldal700}
\end{figure*}

\begin{figure*}[t]
    \centering
    \def\svgwidth{0.975\textwidth}
    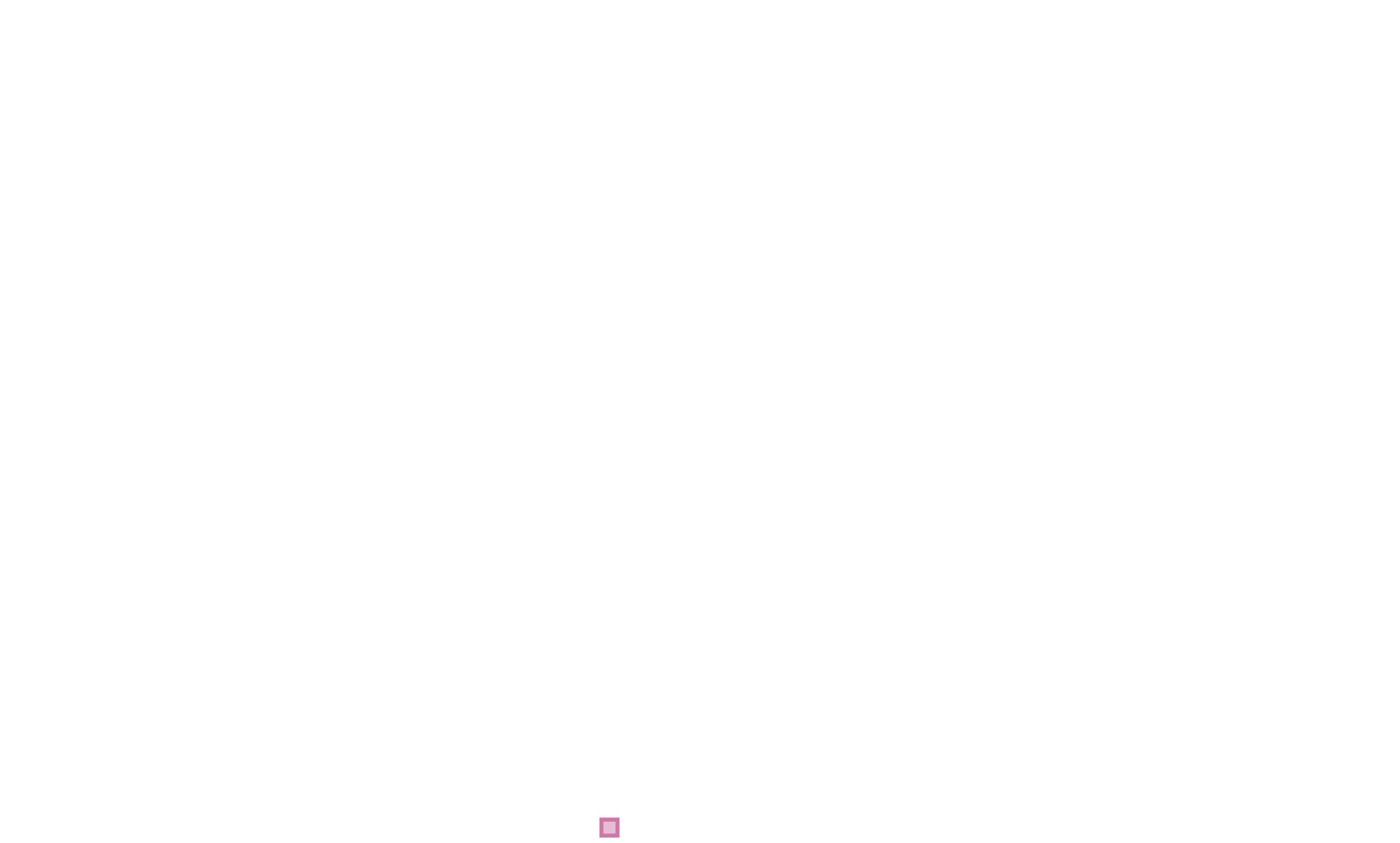
    \caption{\emph{Assumed sensor accuracy estimation errors} when using the BayesCPF on robots with \emph{non-catastrophic sensor degradation ($b[k] \rightarrow 0.7$)}; lower values indicate better performance. Each box (and its outliers) contains data points from $M = 30$ trials. The initial true accuracy $b[0]$ is either $1.0$ (left column, secondary $x$-axis) or $0.8$ (right column, secondary $x$-axis). Although the assumed sensor accuracy estimation here is worse than those in \Cref{fig:delta_cumulative_rmsd_sensor_acc_ldal500}, a higher $\hat{b}[k]$ value in the non-catastrophic case leads to a better informed estimation performance, which we see by comparing \Cref{fig:delta_cumulative_rmsd_regular_inf_est_ldal500} and \Cref{fig:delta_cumulative_rmsd_regular_inf_est_ldal700}.}
    \label{fig:delta_cumulative_rmsd_sensor_acc_ldal700}
\end{figure*}

\subsection{Non-catastrophic Sensor Degradation ($b[k] \rightarrow 0.7$)} \label{sec:results_ldal700}

Following the shortcoming of the BayesCPF in the equilibrium phase, we ask: \textit{If the true sensor accuracy $b[k]$ stabilizes to a level that is not entirely unusable, would estimation performance improve?} This reflects a circumstance where a robot's detection becomes less accurate without being entirely ineffective. This subsection addresses this by simulating a hard bound on the sensor degradation ($b[k] \rightarrow 0.7$). Note that we do not incorporate this knowledge in the BayesCPF, \textit{i.e.,} the assumed sensor accuracy $\hat{b}[k]$ is allowed to fall to 0.5. As shown in \Cref{fig:delta_cumulative_rmsd_regular_inf_est_ldal700}, the answer is undoubtedly positive: we notice a stark improvement in the equilibrium phase on top of a slight N-RMSD decrease in the transient phase.

\paragraph{Transient phase}
Compared to the catastrophic degradation results, the BayesCPF does marginally better in estimating the fill ratio during the transient phase, particularly if the filter is parametrized with the correct assumed sensor accuracy initially ($\hat{b}[0] = b[0]$). The decrease in error is expected since the degradation period is shortened, with the estimation performed using less degraded sensors. Beyond that, when compared to the previous subsection, there are no apparent departures in the N-RMSD trends concerning $f$.

\paragraph{Equilibrium phase}
In this phase, the robots' true sensor accuracy is bounded at $b[k] = 0.7$ (with a $\mathcal{N}(0, \gamma^2)$ noise that occasionally raises it above 0.7). We see a drastic drop in estimation error with respect to the catastrophic degradation case in \Cref{fig:delta_cumulative_rmsd_regular_inf_est_ldal500}. This is consistent with intuition as the number of black tiles observed by the robots' ($n[k]$) is now \emph{not} completely random.

Having a more informative $n[k]$ is also the reason behind the contrasting trends of the N-RMSD between catastrophic and non-catastrophic sensor degradation results. Here, the non-arbitrary $n[k]$ values make the difficulty of estimating the assumed sensor accuracy $\hat{b}[k]$ reliant on the environment fill ratio $f$, as illustrated in \Cref{fig:delta_cumulative_rmsd_sensor_acc_ldal700}. (The only exception is the $\hat{b}[0] - b[0] = +0.2$ case, where we see a more uniform spread of N-RMSD levels, though this is due to the termination of the experiment before the settling of the errors.) The more accurate the assumed sensor accuracy $\hat{b}[k]$, the lower the N-RMSD of the informed estimate $x[k]$. That said, we stress that the \emph{actual accuracy} matters here: \Cref{fig:delta_cumulative_rmsd_sensor_acc_ldal500} shows that while $\hat{b}[k]$ is highly accurate in the equilibrium phase, largely by construction, $b[k] = 0.5$ generates uninformative observations, hence the highly erroneous $x[k]$ in \Cref{fig:delta_cumulative_rmsd_regular_inf_est_ldal500}.

We remark a curious result whereby the N-RMSD \emph{increases with the assumed drift coefficient $\tilde{\eta}$} regardless of correctness. Our view is that this originates from the \emph{constant degradation rate assumption} in the state transition model of the Sensor Accuracy Estimation (SAE) module, which is violated in the equilibrium phase. Since the true sensor accuracy is unchanged (bar minor perturbations), the assumption causes the \textit{predict} step to impose a downward ``pressure'' on $\hat{b}[k]$ that scales with the magnitude of $\tilde{\eta}$. A higher assumed drift coefficient $\tilde{\eta}$ forces the assumed sensor accuracy toward $0.5$ (a singularity in the local estimate computation, \Cref{eq:local_estimate}), causing informed estimation to be more volatile than if $\tilde{\eta}$ is lower. This effect is exacerbated by the environment ambiguity, which---as mentioned earlier---is challenging for sensor accuracy estimation, hence the larger rise in informed estimation errors for environments with a lower $f$. The environmental ambiguity effect even carries over to certain results in the transient phase, where we see a few $x[k]$ N-RMSD levels (\textit{e.g.,} $f = 0.55, 0.65$) scale with the magnitude of $\tilde{\eta}$; assumed sensor accuracy estimation is poor in these cases such that the BayesCPF produces highly volatile informed estimates.

As a whole, we see favorable informed estimation performance with the correct initialization of the BayesCPF like in \Cref{sec:results_ldal500}, but more so here due to the non-catastrophic lower bound. More importantly, the estimation performance significantly improves in the equilibrium phase, which bodes well in applications where sensor degradation is not dire.

\subsection{Ablation Study}

\begin{figure}
    \centering
    \def\svgwidth{\columnwidth}
    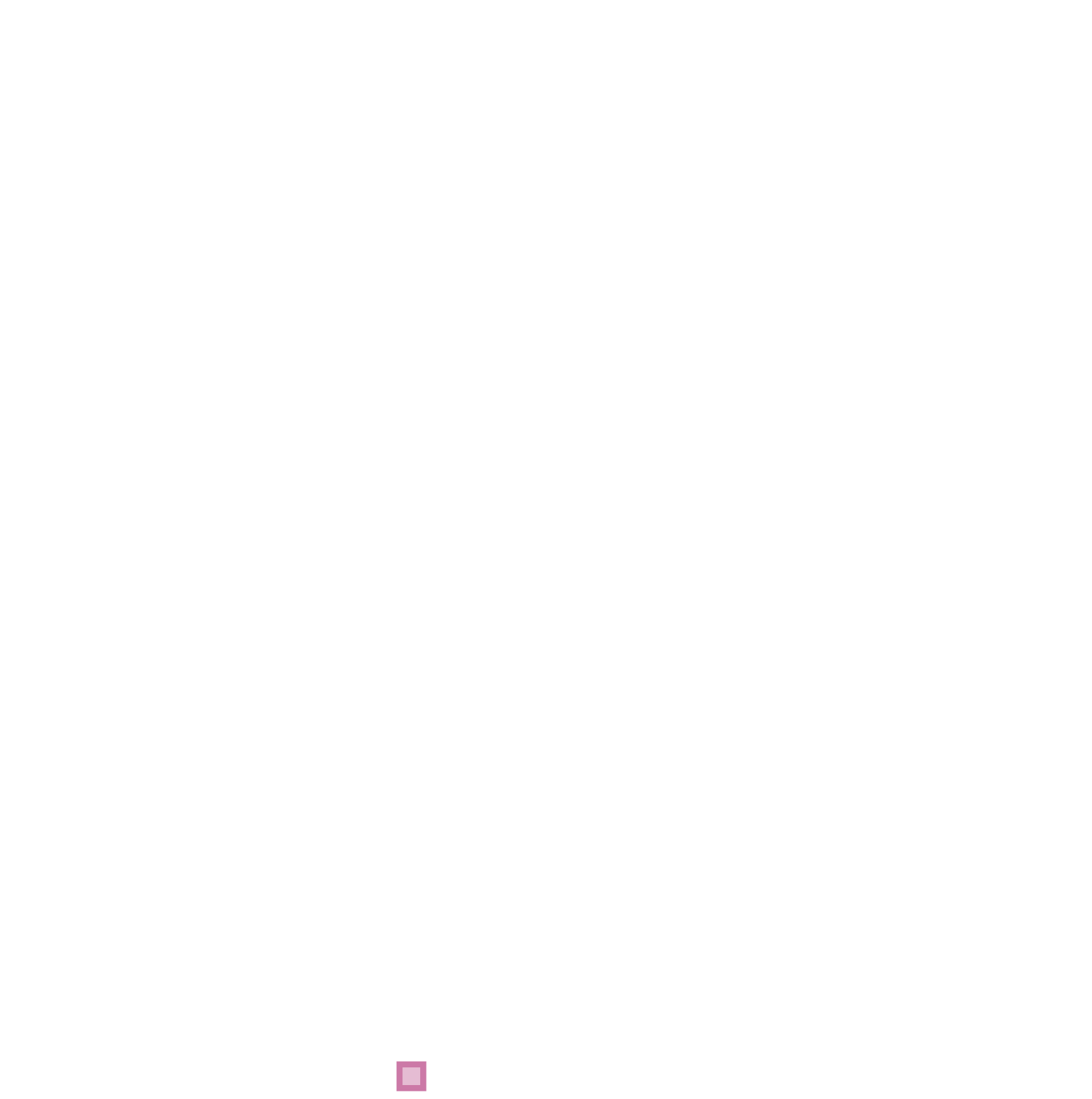
    \caption{\emph{Informed estimation errors} when using the Ablation Filter; lower values indicate better performance. Each box (and its outliers) contains data points from $M = 30$ trials. Note that the rows of this figure describe the case when sensor degradation is catastrophic (top) or non-catastrophic (bottom). The initial true accuracy $b[0]$ is either $1.0$ (left column, secondary $x$-axis) or $0.8$ (right column, secondary $x$-axis). The singularity at $b = 0.5$ introduced by the MCP algorithm (\Cref{eq:local_estimate}) precludes accurate informed estimation during the equilibrium phase, even when complete sensor accuracy information is available.}
    \label{fig:oracle_cumulative_rmsd_regular_inf_est}
\end{figure}

As the Fill Ratio Estimation (FRE) module (\Cref{sec:fre_module}) is derived from previous work, the core contribution of this work is in the estimation of a dynamically degrading sensor, \textit{i.e.,} the SAE module---and the auxiliary Constraint Compliance (CC), Observation Queue Adjustment (OQA) modules---in \Cref{sec:sae_module,sec:cc_module,sec:oqa_module}. This subsection assesses the SAE module's impact by removing the sensor degradation estimation functionality of the BayesCPF. The robots have complete information on their true sensor accuracy, which is used in the OQA module and the FRE modules: $\hat{b}[k] \coloneq b[k]$. We refer to this reduced algorithm as the \emph{Ablation Filter (AF)}. For the most part, \Cref{fig:oracle_cumulative_rmsd_regular_inf_est} shows that the AF has good estimation performance in situations where information can be salvaged, \textit{i.e.,} the transient phase of the catastrophic degradation experiments and both phases of the non-catastrophic degradation experiments.

We introduce a metric called the \emph{N-RMSD Difference} to compare the performance of the BayesCPF and the AF:
\begin{align}
    \Delta \zeta_x^{[m]} & =  \zeta_x^{[m]} - \zeta_x^{\star [m]}
\end{align}
where $\zeta_x^{\star [m]}$ is the informed estimation N-RMSD of an experiment trial $m$ obtained with the AF. The performance of the BayesCPF is superior to the AF when $\Delta \zeta_x^{[m]} < 0$. (We only analyze the N-RMSD difference of $x[k]$ because the AF has no assumed sensor accuracy errors, \textit{i.e.,} $\Delta \zeta_b^{[m]} = \zeta_b^{[m]}$.) The conventional expectation is that the AF should surpass the BayesCPF in estimation performance, given that the former's information on sensor accuracy is perfect; nonetheless, the outcome is not straightforward.

\begin{figure*}
    \centering
    \begin{subfigure}{\textwidth}
        \def\svgwidth{0.97\textwidth}
        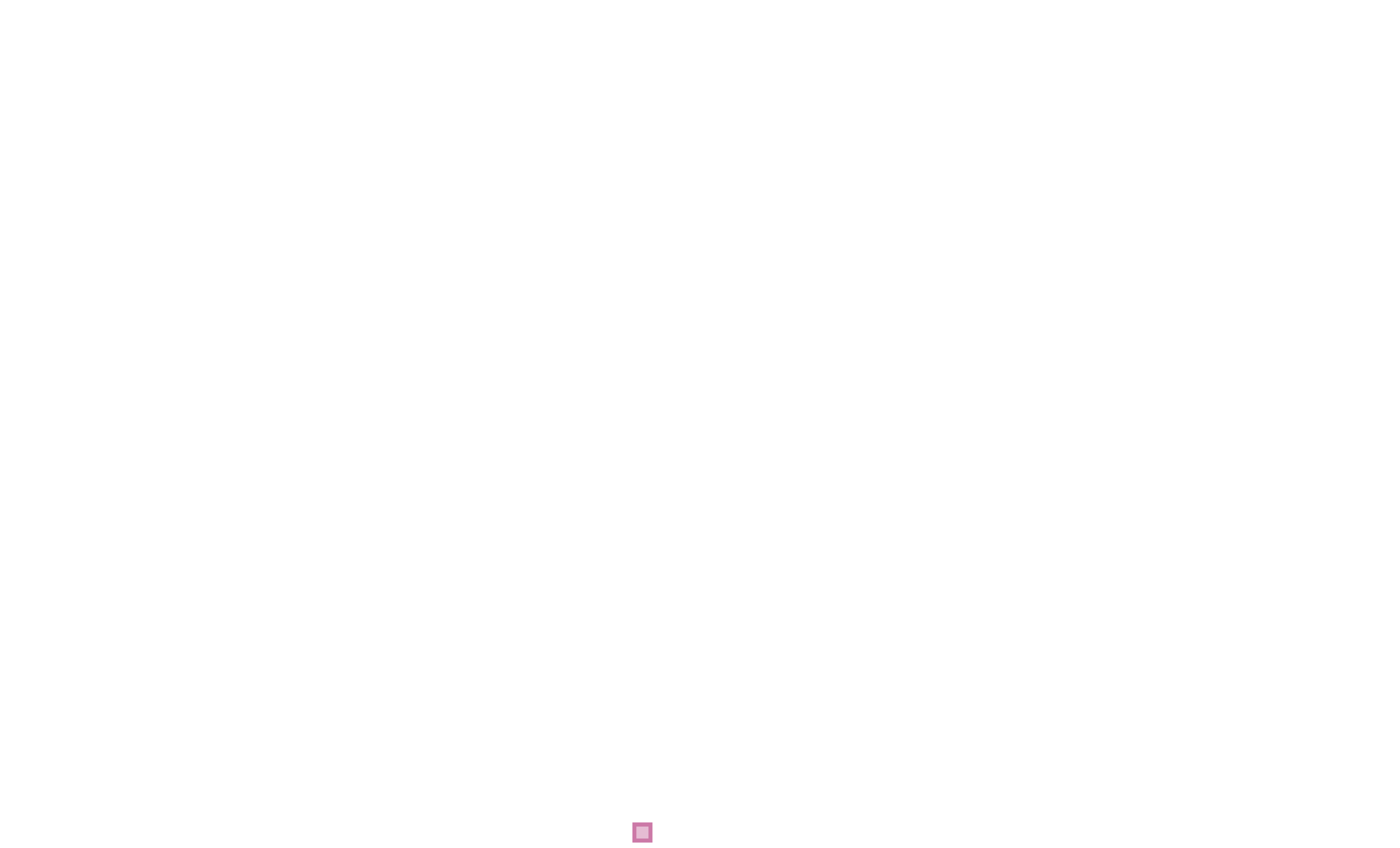
        \caption{\emph{Catastrophic sensor degradation} ($b[k] \rightarrow 0.5$).}
        \label{fig:perf_ind_regular_inf_est_ldal500}
    \end{subfigure}
    \begin{subfigure}{\textwidth}
        \def\svgwidth{0.97\textwidth}
        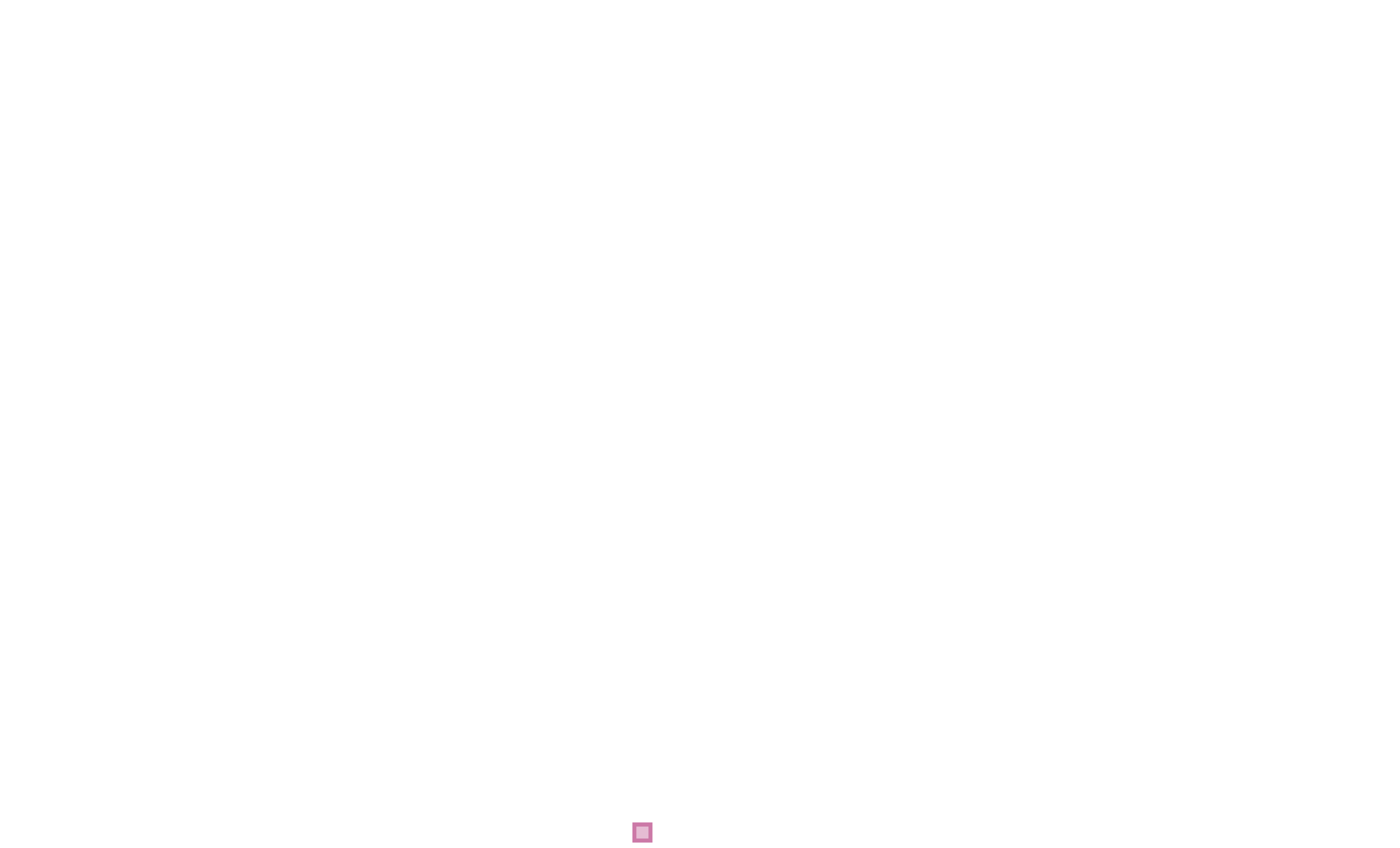
        \caption{\emph{Non-catastrophic sensor degradation} ($b[k] \rightarrow 0.7$).}
        \label{fig:perf_ind_regular_inf_est_ldal700}
    \end{subfigure}
    \caption{\emph{Difference in informed estimation errors} between the BayesCPF and the AF; $\Delta \zeta_x \leq 0$ indicates performance by the BayesCPF that is on par with the AF or better. Each box (and its outliers) contains data points from $M = 30$ trials. The initial true accuracy $b[0]$ is either $1.0$ (left column, secondary $x$-axis) or $0.8$ (right column, secondary $x$-axis). The performance edge by the BayesCPF---when sensor degradation is catastrophic---is due to the unpredictability of using an unusable sensor.}
    \label{fig:perf_ind_regular_inf_est}
\end{figure*}

\subsubsection{Catastrophic Sensor Degradation ($b[k] \rightarrow 0.5$)}
Due to the poor assumed sensor accuracy estimation by the BayesCPF, \Cref{fig:perf_ind_regular_inf_est_ldal500} shows the superior estimation performance of the AF in the transient phase, as anticipated. Interestingly, we see a slight edge by the BayesCPF in an ambiguous environment ($f = 0.55$) when the \emph{degradation model is underestimated} ($|\tilde{\eta}| < |\eta|$) and the \emph{initial assumed sensor accuracy is overestimated} ($\hat{b}[0] - b[0] = +0.2$). This phenomenon stems from the manifestation of the volatility in the black tile observation rate $\nicefrac{n[k]}{t[k]}$. At $f = 0.55$ in \Cref{fig:heatmap} (left), $\nicefrac{n[k]}{t[k]} \approx 0.55$ can be produced by any true sensor accuracy ($0.5 \leq b[k] \leq 1.0$), bearing in mind that observations are stochastic.

In the SAE module of the BayesCPF, the volatility in $\nicefrac{n[k]}{t[k]}$ reflects on the assumed sensor accuracy $\hat{b}[k]$ because its estimation relies on the WMA informed estimate $x^\prime[k]$ as the reference to $f$. In other words, the reference is the ``authoritative source'' of information and is thus constant; consequently, any changes in $\nicefrac{n[k]}{t[k]}$ carry directly over to $\hat{b}[k]$. For the AF, $\hat{b}[k]$ is fixed to $b[k]$, which means that the noise in $\nicefrac{n[k]}{t[k]}$ is instead reflected in $x[k]$. As a result, the higher informed estimate volatility in the AF leads to a worse N-RMSD when compared to the BayesCPF.

Another surprising outcome where the BayesCPF emerges superior over the AF is found in the equilibrium phase for every configuration we tested. Here, the AF computes $\hat{x}[k]$ based on $\hat{b}[k]$ that is close to 0.5 (rightly so) but with variance orders of magnitude lower than the BayesCPF. We see this by comparing \Cref{fig:true_sensor_acc_example} against the assumed sensor accuracy plots in \Cref{fig:time_series_ldal500}: $b[k]$ stays close to 0.5 with significantly smaller perturbations than $\hat{b}[k]$. Effectively, similar to the transient phase, the steadier $\hat{b}[k]$ used by the AF leads to an erratic $\hat{x}[k]$ that oscillates between 1.0 and 0.0. The BayesCPF, on the other hand, benefits from the fortuitous outcome where the more volatile $\hat{b}[k]$ generates $\hat{x}[k]$ values that are $> 0.0$ and $< 1.0$ which, while still highly erroneous, is \emph{more accurate} than the AF estimates.

So far, besides a few experimental configurations in the transient phase, the BayesCPF unexpectedly stands out against the AF, particularly in the equilibrium phase. This advantage, however, does not come explicitly from the BayesCPF but is instead due to \Cref{eq:local_estimate}. Despite this, with correct initialization of the BayesCPF, we can still obtain performance similar to the case where sensor accuracy is known perfectly (while the sensor is degrading).

\subsubsection{Non-catastrophic Sensor Degradation ($b[k] \rightarrow 0.7$)}
Here, the N-RMSD difference between the BayesCPF and the AF falls in line with expectation: $x[k]$ estimation is generally better with perfect sensor accuracy information (\Cref{fig:perf_ind_regular_inf_est_ldal700}). Still, we note that the BayesCPF produces performance comparable to the AF in the transient phase when correct initial conditions are provided regardless of $\tilde{\eta}$, a positive sign for our approach. With the lack of perfect sensor accuracy information, additional difficulties in the experimental configuration (\textit{e.g.,} incorrect initial conditions, degradation model mismatch in the equilibrium phase) make the estimation problem more challenging, hence the inferior BayesCPF performance.

In the transient phase, aside from when the initial conditions are correct, there are certain instances where the estimation errors of the BayesCPF are close to the AF. When initial conditions are underestimated, the almost negligible error difference for the $f = 0.95$ case can be attributed to the accurate $\hat{b}[k]$ estimation by the BayesCPF. Meanwhile, when initial conditions are overestimated, the similar N-RMSD levels in the $f = 0.55$ are due to the vicinity of the solution for $\hat{x}[k]$ (by the BayesCPF) to the ground truth regardless of the true accuracy, as mentioned previously in \Cref{sec:results_ldal500}.

In the equilibrium phase, we see the role of $\hat{b}[k]$ estimation performance in the differences in N-RMSD levels. The trends of the N-RMSD difference (of $x[k]$) follow the trends of the N-RMSD levels of the BayesCPF (\Cref{fig:delta_cumulative_rmsd_regular_inf_est_ldal700,fig:delta_cumulative_rmsd_sensor_acc_ldal700}). (Recall that the inverted trend for the case with overestimated initial conditions only appears so because the experiments have been terminated before the error levels have settled.) This fits our expectation that better sensor accuracy estimation yields better-informed estimation.

Overall, the AF results show that the quality of sensor accuracy estimation influences the accuracy of the informed estimation, and that the BayesCPF is competitive in situations when correctly initialized. At the same time, the BayesCPF benefits from fortuitous outcomes---largely by construction---that see it outperform circumstances where sensor accuracy levels reach the lowest point.
\section{Physical Experiments} \label{sec:physical_experiments}

\begin{figure*}[t]
    \centering
    \def\svgwidth{0.975\textwidth}
    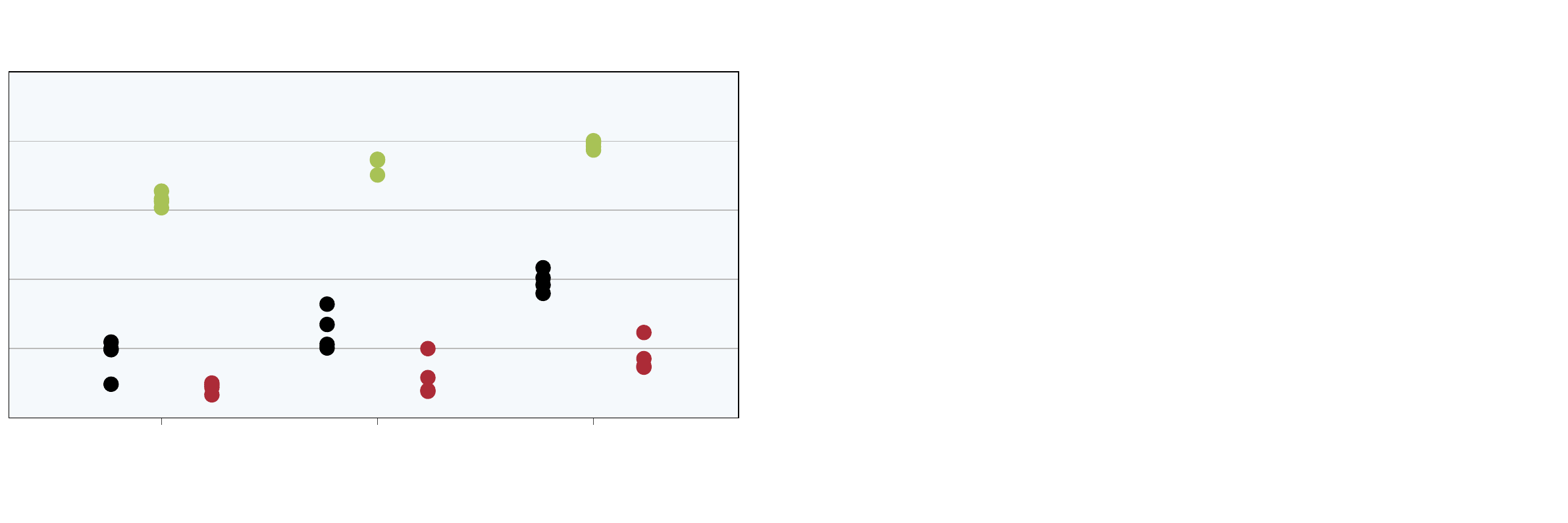
    \caption{\emph{Informed estimation errors} from physical robot experiments where robots have a \emph{non-catastrophic sensor degradation ($b[k] \rightarrow 0.7$)}; lower values indicate better performance. Each data point represents the N-RMSD of one trial (out of four) for each experimental configuration. Because we only test for one fill ratio, $f = 0.448$, and run fewer trials, the current box plot layout differs from the simulated experiments shown earlier. We group the different $\tilde{\eta}$ parameters in separate $x$-axis columns and use the colors to indicate the different initial conditions.}
    \label{fig:physical_delta_cumulative_rmsd_regular_inf_est_ldal700}
\end{figure*}

\subsection{General Setup}
Our physical experiments use six Khepera IV robots, like those used in the simulated experiments. They are equipped with a Texas Instruments DM3730 800MHz ARM Cortex-A8 processor with 512 MB of memory and \unit[4]{GB} of storage space for us to run the BayesCPF onboard. For communication, the robots transmit multicast UDP messages over a shared Wi-Fi network. Our communication scheme is partially decentralized, as each robot broadcasts messages for its neighbors to receive, with a Wi-Fi router facilitating message propagation as the network backbone. We simulate a single-hop message broadcasting range for a robot by filtering out messages from neighbors \emph{outside} the single-hop range $R = \unit[0.7]{m}$. A Vicon motion capture system provides the localization information that we use to do so. Like in the simulated experiments, a robot diffuses across an environment while avoiding obstacles, which is done using its eight proximity IR sensors. The ground sensor is also an IR sensor that detects whether a tile is black. We run each of the four trials per experiment for $K_{max} = 20{,}000$ steps (each time step equivalent to $\unit[0.1]{s}$), totaling more than 22 hours of physical experiment data in this work.

The \emph{true drift coefficient} is simulated at $\eta = -3 \times 10^{-5}$ while we test the BayesCPF with \emph{assumed} drift coefficients of $\tilde{\eta} = \{-1.5, -3.0, -4.5\} \times 10^{-5}$. We do this in the robot controller by modifying its actual observation akin to flipping a $b[k]$-weighted coin: if a randomly generated number (between 0.0 and 1.0) is $< b[k]$, then the observation is assigned---without changes---to $z[k]$, to be used by the BayesCPF. Otherwise, the opposite color is assigned instead. We consider the following initial sensor accuracy conditions in our physical experiments.
\begin{itemize}
    \item Correct: $\hat{b}[0]$ = $b[0]$ = $1.0$
    \item Incorrect (underestimated): $\hat{b}[0]$ - $b[0]$ = 0.8 - 1.0 = -0.2
    \item Incorrect (overestimated): $\hat{b}[0]$ - $b[0]$ = 1.0 - 0.8 = +0.2
\end{itemize}
The robots diffuse across a $3.0353 \times \unit[3.0226]{m^2}$ QR code arena with a fill ratio of $f = 0.448$ (bottom right of \Cref{fig:bayes_cpf_block_diagram}), moving at $\unit[7]{cm / s}$. They observe the environment and run the BayesCPF at a rate of $T_{obs} = T_{BCPF} = \unit[5]{Hz}$ and communicate at a rate of $T_{comms} = \unit[2]{Hz}$. Besides that, all the experimental and BayesCPF parameters remain as listed in the simulated experiment section.

\subsection{Results and Discussion}
To better contextualize the physical experiment results shown in \Cref{fig:physical_delta_cumulative_rmsd_regular_inf_est_ldal700}, we compare them against those of the simulated experiments in \Cref{fig:delta_cumulative_rmsd_regular_inf_est_ldal700}, where the environment fill ratio is $f = 0.55$. We do this because the environments with $f = 0.448 \approx 0.45$ (physical) and $f = 0.55$ (simulated) are highly similar from a binary estimation perspective; the problem is symmetrical at $f = 0.5$. Additionally, we provide a benchmark result---marked by the horizontal gray dashed line in \Cref{fig:physical_delta_cumulative_rmsd_regular_inf_est_ldal700}---where the robots have a non-degrading, perfect sensor and are aware of its sensor accuracy ($\hat{b}[k] = b[k] = 1.0$).

The overall informed estimation trends here (with respect to the initial conditions) match the simulated experiment results: estimation performance is more sensitive to initial conditions than the correctness of degradation models. As expected, performance in the transient phase is better than in the equilibrium phase. Likewise, the N-RMSD levels are the highest for the $\hat{b}[0]$ - $b[0]$ = -0.2 category as compared to the cases where $\hat{b}[0]$ = $b[0]$ = 1.0 or $\hat{b}[0]$ - $b[0]$ = +0.2. We also see that the estimation performance rivals that of the perfect sensing case (horizontal gray dashed line) in the transient sensor degradation region for certain cases (where $|\tilde{\eta}| < |\eta|$, $\hat{b}[0] = b[0]$ and $\hat{b}[0] > b[0]$).

Different from the simulated experiments, however, the physical experiment's informed estimation errors are noticeably higher. Comparing case-by-case, we see that the N-RMSD levels of physical robot experiments are elevated compared to their simulated counterparts. Several reasons cause this. First, the true sensor degradation rate $\eta$ is $3 \times$ higher here than it is in the simulated experiments. We attempted to compensate for that by increasing the observation rate $\nicefrac{1}{T_{obs}}$ (and, consequently, the activation rate for the SAE, CC, and OQA modules $\nicefrac{1}{T_{BCPF}}$). However, it appears that using the simulated experiment values for $\phi$, $W$, and $Q_{max}$---``dials'' used in the OQA module (\Cref{sec:oqa_module}) to adjust for the filter's sensitivity to the change in the observations' distribution---is too permissive for the increased degradation rate. Second, on top of a smaller swarm size of six robots, communication between robots can be challenging as corrupted messages are sometimes received (which are discarded); this is not a problem for the simulated robots. The real robot controller is also compiled to use 32-bit \texttt{float} numbers, whereas the simulated controller uses 64-bit \texttt{double} numbers, leading to minor computation discrepancies between the two. Third, QR code cells are not uniformly distributed, even though they appear largely \emph{well-mixed}---such that a Bernoulli process can model the observations well. This means that the binomial likelihood formulated by the MCP algorithm (used in the FRE module in \Cref{sec:fre_module}) is no longer appropriate. While the violation of the binomial model by the QR code does not appear severe, it is nonetheless a probable cause for the worse estimation performance. We believe that these reasons, alongside the highly challenging environment fill ratio of $f = 0.448$ (as explained in \Cref{sec:simulated_experiments}), are responsible for the worse performance in the physical experiment results.

While these reasons account for the elevated N-RMSD levels across the board, we suspect that the main culprit behind the drastic error increase for the $\hat{b}[0]$ - $b[0]$ = -0.2 case to be the \emph{starting point of the initial assumed sensor accuracy}. In every physical experiment, the robots' assumed sensor accuracy eventually goes to $\hat{b}k$ = 0.5 despite non-catastrophic sensor degradation ($b[k] \rightarrow 0.7$)---this is as expected due to the ambiguous environment, as discussed in the simulated experiment section. Unlike the cases where the initial assumed accuracy is $\hat{b}[0]$ = 1.0, however, starting with a lower $\hat{b}[0]$ means that the sensor accuracy estimate approaches catastrophic degradation ($\hat{b}[k] = 0.5$) faster than it would with a higher $\hat{b}[0]$. This is exacerbated by the state transition model, which---as explained earlier---induces a downward pressure on $\hat{b}[k]$. All in all, we surmise that the combination of all these factors is what drives the significant increase in error for the $\hat{b}[0]$ - $b[0]$ = -0.2 case of our physical experiments.

The key takeaway from our physical experiment results is that the BayesCPF performs well in computationally constrained systems, particularly when initial conditions are known. We showed that onboard these robots, estimation performance is competitive to the case when robots have perfect sensing (and are aware of it) if initial assumed accuracies are calibrated. Our use of the Khepera IV robot, which fits the profile of a typical swarm robot, demonstrates the viability of the BayesCPF in computationally limited robots. Furthermore, for more capable robots, the BayesCPF can supplement advanced estimation algorithms that may be more computationally expensive.
\section{Discussion and Conclusion} \label{sec:conclusion}

In this paper, we presented the Bayes Collective Perception Filter (BayesCPF), an approach to collective perception using swarm robots with degrading sensors. The BayesCPF enables a robot to simultaneously estimate its environment fill ratio and sensor accuracy level by implementing an Extended Kalman Filter and the Minimalistic Collective Perception algorithm.

An important quality of the BayesCPF is that it applies to decentralized robot swarms, with each individual robot running its own copy of the filter. Although the problem setup we used is ostensibly a simplified one (\Cref{fig:mcp_setup})---such that a single-robot solution may suffice---a multi-robot solution offers crucial advantages. First, a wider coverage area can be surveyed \emph{faster} by a group of robots than a single individual. Then, the authors of the MCP algorithm highlighted the importance of communication among these robots, resulting in faster convergence of the estimate. Next, we argue that the BayesCPF produces \emph{less volatile estimates} in the presence of neighboring robots, since it relies on a weighted moving average of informed estimates---which combines neighboring estimates---as a reference for the fill ratio. This contributes further to faster convergence of the robot swarm's estimate. Hence, the BayesCPF can capitalize on the benefits of a multi-robot system while being flexible enough for a single-robot solution.

While the derivation demonstrated in this work applies to a binary problem, it can be generalized to higher-dimensional problems, though specific problems are easier than others. For example, if the observation outcomes are \emph{mutually exclusive}---like when estimating if the ambient temperature is cold, suitable, or hot---a multinomial distribution can be used in the FRE module since its maximum likelihood estimator is well-defined. Similarly, the EKF is designed to handle multidimensional states, making generalizing the SAE module straightforward. Efforts to generalize the CC module are also minimal, as the PDF truncation method from \cite{simonConstrainedKalmanFiltering2010} was originally designed with multidimensional state variables in mind. The OQA module is unchanged, except that multiple copies tracking the different sensor accuracies will need to be run simultaneously. It becomes non-trivial, however, to extend our proposed method if multiple outcomes can be simultaneously observed, as that changes the maximum likelihood estimator in the FRE module. A significant rework may be required of the FRE module, which may lead to the modification of the other modules.

From our simulated and physical experiments, we identified the strengths and weaknesses of the BayesCPF. In conditions where the sensor degrades unbounded, the BayesCPF can provide usable estimates when the degradation assumption holds---\textit{i.e.,} the transient degradation phase---but not so beyond that; no useful information can be extracted from a completely broken sensor to be used in generating coherent fill ratio estimates. As such, we view the poor estimation performance in the equilibrium phase as a natural consequence rather than a reflection of our approach's deficiency. This is further corroborated by an ablation study in which a filter equipped with complete sensor accuracy information---the Ablation Filter (AF)---is compared with the BayesCPF. The comparison shows unexpected estimation superiority by the BayesCPF because of the unpredictable observations when sensor degradation is catastrophic.

When sensor degradation is not catastrophic, we still see competitive estimation performance by the BayesCPF, even as compared to the AF. This holds in the transient phase where the appropriate sensor accuracy level has been provided initially, regardless of the modeling choice. Otherwise, the BayesCPF is limited when the degradation model is violated during the equilibrium phase, though more so for \emph{ambiguous environments} than \emph{unambiguous ones}. Using the Khepera IV robots that only communicate locally, our physical experimentation demonstrates the feasibility of the BayesCPF as a minimal algorithm that can compensate for dynamic sensor degradation for collective perception. Moreover, the algorithm's small computational and memory footprint makes it an attractive option to supplement estimation algorithms that are more computationally intensive.

In the future, we intend to explore sensor degradation profiles whose expectation is nonmonotonic or nonlinear, as they cover a wider scope of detection failures that we see in the real world. Additionally, we are interested in the problem of monitoring a dynamically changing environment---like those studied by Pfister \textit{et al.}~\cite{pfisterCollectiveDecisionmakingBayesian2022,pfisterCollectiveDecisionmakingChange2023}---using degrading sensors, as there exist many real-world examples with transitory elements (\textit{e.g.,} air pollution, microplastic transport, etc.).

\balance

\bibliographystyle{IEEEtran}
\bibliography{IEEEabrv,references}

\end{document}